\documentclass[fleqn,10pt]{wlscirep}  
\usepackage{xr} 

\usepackage[utf8]{inputenc}
\usepackage{units}
\usepackage{mathtools}  
\usepackage{amsmath}
\usepackage{amssymb}
\usepackage[dvipsnames]{xcolor}
\usepackage{commath}  
\usepackage{subcaption}
\usepackage{comment}
\usepackage[T1]{fontenc}
\title{Self-governed access control in open robot swarms using a blockchain-based DAO}

\author[1,*]{Raina Zakir}
\author[1]{Alexandre Pacheco}
\author[2,3]{Andreagiovanni Reina}
\author[1]{Marco Dorigo}
\author[1,*]{Volker Strobel}

\affil[1]{IRIDIA, Université Libre de Bruxelles, Brussels, Belgium}
\affil[2]{Centre for the Advanced Study of Collective Behaviour, Universit\"{a}t Konstanz, Konstanz, Germany}
\affil[3]{Department of Collective Behaviour, Max Planck Institute of Animal Behavior, Konstanz, Germany}

\affil[*]{raina.zakir@ulb.be; volker.strobel@ulb.be}

\newif\ifcmnts
\cmntstrue

\begin{abstract}

Robot swarms harness simple, local interactions among many individually limited robots to produce complex, coordinated collective behavior, making them valuable across domains from agriculture to disaster response. We envision the next frontier for such swarms as open swarms, in which robots join or leave the swarm dynamically at runtime. The ability to dynamically adjust swarm size has been shown to be beneficial, for instance by reducing inconsistencies and network partitioning, especially under sparse connectivity or in complex environments, ultimately improving task performance. For practical deployment, however, open swarms requires governance, particularly protection against Byzantine behavior (i.e., faulty or malicious robots), since allowing robots to join freely also makes the swarm easier to infiltrate. Blockchain technology has recently been incorporated into robot swarms, offering secure, decentralized coordination and improved resilience, though largely in closed settings with a fixed, predetermined membership. In this work, we extend these benefits to open swarms by introducing robot swarms governed by a blockchain-based decentralized autonomous organization (DAO), in which the swarm regulates itself using autonomous rules encoded in smart contracts. We outline several DAO-based governance mechanisms for robot swarms, depending on the needs of the application, and present a proof-of-concept implementation of a simple DAO that regulates swarm membership and prevents Byzantine infiltration. Our simulation results and real-robot validations show that DAO-governed access control and robot removal can maintain proper swarm functioning while keeping the number of Byzantine robots within acceptable limits.

\end{abstract}
\begin{document}

\flushbottom
\maketitle

\thispagestyle{empty}

\section*{Introduction}

Robot swarms are decentralized systems, typically composed of a large number of autonomous robots that coordinate to perform tasks~\cite{DorBirBra2014:sch-sr,DorTheTri2021:pieee,Hamann2018}, achieving outcomes that no single robot could accomplish alone. Within such systems, robots exchange information and task-critical data with one another. These local interactions, both among robots and with their environment, give rise to self-organized collective behavior. Swarm robotics systems have been shown to hold potential across a wide range of domains, including agriculture~\cite{albani_agri_2017,albiero2022Agri}, environmental monitoring~\cite{tarapore2020sparse,huang_explore_2019}, construction~\cite{werfel_construct_2014,allwright2017}, and disaster response operations~\cite{Tzoumas2024SI,shi2025wildfires}. The decentralized nature of robot swarms facilitates parallel task execution and ensures scalability across swarm sizes, while also providing fault tolerance against individual failures. Realizing these benefits, however, depends on effective coordination among robots, each of which must act on consistent information and pursue shared goals across the swarm. To date, such coordination has largely been achieved within swarms deployed and owned by a single entity, in which membership is fixed and known in advance. As robotics technology matures and its use across real-world applications continues to gain traction, we envision \textit{open swarms} as the new paradigm for robot swarms, i.e., multi-robot teams in which robots join and leave dynamically at runtime. 

The dynamic regulation of swarm size in open swarms brings forth several advantages. When task demand increases, additional robots can be recruited to scale up resources, and when robots become redundant, they can be removed from the swarm or relocated elsewhere to maintain resource efficiency, providing performance benefits over static deployments~\cite{Fuady2025SubCDMCD}. 
Looking ahead, the increasing accessibility of the technology is likely to further drive the deployment of autonomous robots by companies and private users alike, for widespread tasks such as environmental monitoring, waste management, pest control, surveillance, and mapping. As such deployments become more widespread, robots belonging to various stakeholders will increasingly need to operate within shared physical spaces~\cite{DorPacReiStr2024:review,winfield2025ethical}, giving rise to the need for robot swarms that evolve dynamically rather than being operated as a fixed, closed system. 
Dynamic membership, however, introduces a governance challenge. Due to the decentralized nature of robot swarms, no single entity exists to make decisions such as regulating swarm membership, assigning roles, managing hierarchical relationships among robots, upgrading protocols, or allocating resources in an open swarm setting. Among these governance needs, regulating swarm membership, that is, establishing who may join, remain in, or be removed from the swarm, is especially important. As there are no inherent mechanisms to verify their trustworthiness before admission, allowing robots to join freely in open swarms increases the risk of infiltration by Byzantine robots (that is faulty or malicious robots whose behavior deviates from their intended specification), turning this governance gap directly into a risk to task execution. This gap also leads to security concerns, as open membership can expose robotic systems to \emph{Sybil attacks}, in which an adversary creates multiple identities to amplify its influence.




Several studies have highlighted need for security in swarm robotics~\cite{higgins2009threats,HigTomMar2009:icas,HunHau2020:nmi}; yet, as is often the case with emerging technologies, security has largely been treated as an afterthought, and research in this area remains in its infancy. Most studies either consider closed swarms, which restrict participation to a predefined set of identities, or assume non-malicious interactions. In the real-world, open swarms require secure mechanisms to govern membership. 

Research on securing robot swarms has primarily focused on collective decision-making, especially the best-of-$n$ problem, in which a swarm must reach a consensus on the most suitable of $n$ options~\cite{ValFerDor2017:frontiers,ValHamDor2015:aamas,ReiValFer-etal2015:plosone}. In the best-of-$n$ problem, Byzantine behavior, arising from malicious agents or communication manipulation, can lead to consensus deadlocks (analogous to denial-of-service in computing security) or cause the swarm to converge on suboptimal options (known as the wrong addressing problem, since the swarm is led to incorrect choices)~\cite{masi2021robot,antonic2024collective,zakir2026natcomm}. Prior research has proposed defenses such as neighborhood filtering and probabilistic trust models to assess received information based on its consistency with a robot's own data~\cite{Zakir:IROS:2024,ZakDorRei2022:ants,reina2023cross,crosscombe2017iros,moussa2020robustness,PriTucTagFer2018:si,TarChrTim2017:plos}. Other approaches protect the confidentiality and integrity of inter-robot data exchanges while preventing the disclosure of the details of the mission performed by the robot swarm~\cite{CasHarPenDor2021:sciencerobotics,QueraltaIoT2022}. However, these works primarily focus on limiting the impact of Byzantine behavior. More recently, blockchain-based approaches have shown promise for actively identifying and mitigating Byzantine robots~\cite{StrPacDor2023:sciencerobotics,ZhaPacStr-etal2023:iros,MoroncelliAnts2024,VanPacStr-etal2023:SciRep}. Blockchain enables tamper-proof algorithms to be executed in a decentralized manner through smart contracts (see Section~\nameref{methods}, \nameref{blockchain}), supporting not only the control and coordination of swarm behavior but also the active identification and removal of Byzantine robots. However, these implementations assume closed settings in which the total number of robots, and therefore the number of Byzantine robots to be neutralized, is fixed and known a priori, preventing scalability. Recent work has shown that runtime scalability achieved by forming open swarms can effectively regulate latency in blockchain-enabled robot swarms~\cite{zak_ants26}, but this scalability gives rise to a related security concern, since once the swarm is open, the risk of Byzantine infiltration becomes a threat. Similarly, prior work on open membership in general multi-robot systems does not resolve this risk, remaining vulnerable to Sybil attacks or often relying on a centralized authority to vet newly joined members~\cite{BijRob2014:airev,ZikMasLeb-etal2016:rusmart}, an assumption that defeats the vision of a truly open, decentralized swarm. Robot swarms therefore demand a fundamentally different approach, one that can detect and remove Byzantine robots without relying on any centralized control, even as membership changes at runtime.

\begin{figure}[!ht]
\centerline{\includegraphics[width=1\textwidth,trim={0.3cm 20cm 0cm 1cm}]{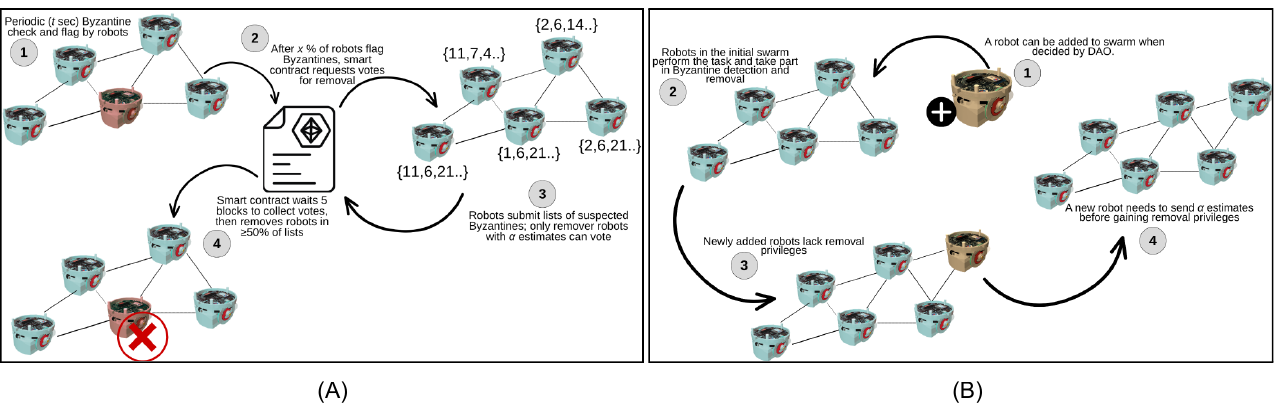}}
\caption{\textbf{DAO implementation.} We present a basic proof-of-concept DAO implemented in a robot swarm. (A) Every $t$ timesteps, each robot checks for Byzantine behavior in the swarm. `When' a robot initiates removal is determined by a threshold-based Byzantine detection mechanism. Once a robot detects a Byzantine, it signals the smart contract, which triggers a voting round. Robots then decide `who' to remove using a non-tokenized, reputation-based mechanism where each robot submits a list of suspected Byzantines to the smart contract. The contract aggregates these reports, and `how' a removal occurs is governed by the rule whereby a robot is removed if it appears on at least 50\% of the submitted lists after a 5-block waiting period. (B)~Our DAO operates effectively as long as the influx of robots does not exceed the system's ability to remove Byzantine robots. For every robot removed, one or more new robots join the swarm. These new robots initially do not have the right to vote on the removal of other robots. The parameter $\alpha$ modulates how long newly added robots must provide valid estimates before they are allowed to participate in the removal decision.}
\label{fig:daoimp}
\end{figure}

\begin{figure}[!ht]
\centerline{\includegraphics[width=1\textwidth,trim={0cm 12cm 0cm 2cm}]{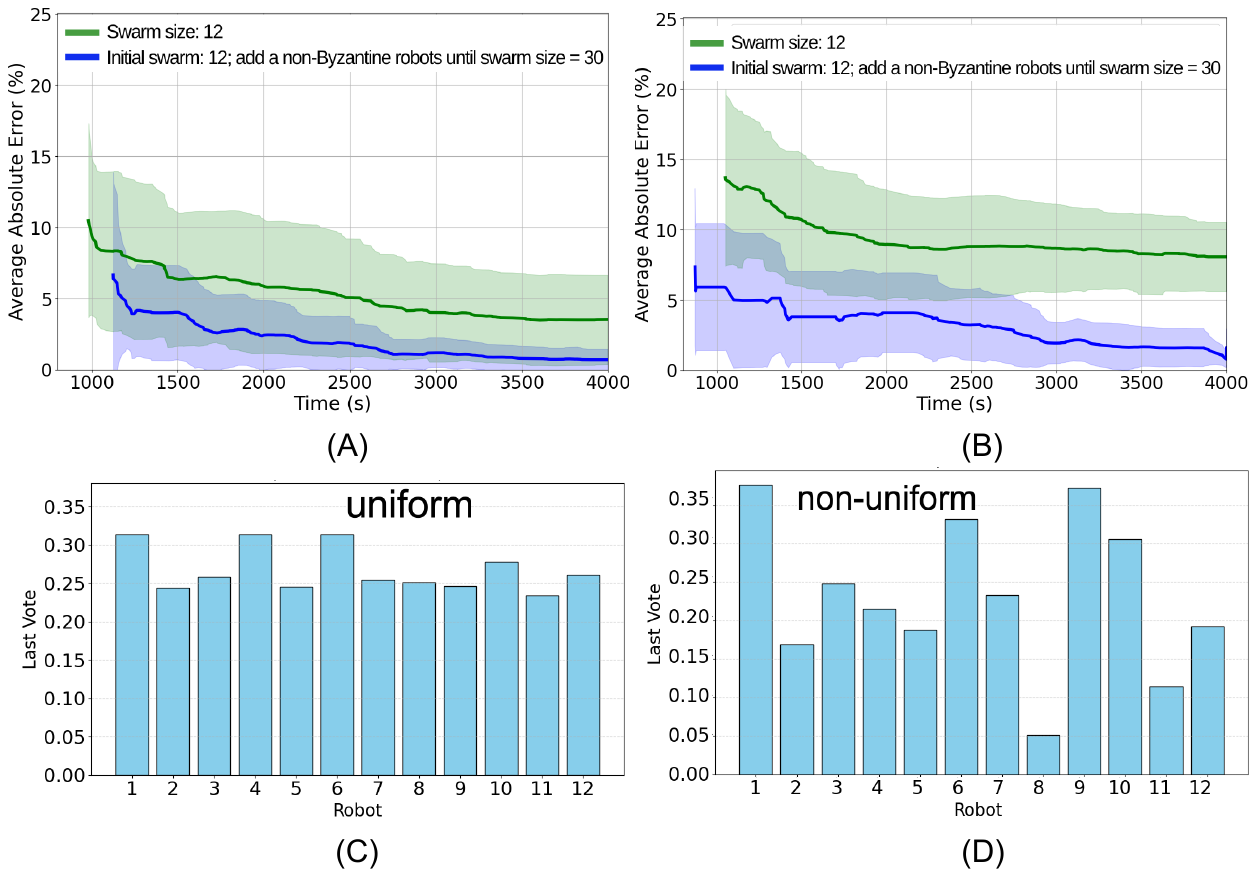}}
\caption{\textbf{Motivation for open swarms}. 
(A–B) Average absolute error over time from the true 25\,\% value across 10~runs, with standard deviation (showing variability across runs). To demonstrate the need for open swarms, we consider a larger arena of $15.40$m$^2$ (instead of $3.55$m$^2$) for the results presented in this figure. (A) A static swarm of 12 robots (green) struggles, whereas adding honest robots every 120\,s until the swarm reaches 30~robots (blue) drives the error toward zero. (B) Similarly, when the initial swarm has 33\,\% Byzantines, adding honest robots reduces Byzantine influence. The experiments in (A-B) are run without the outlier-based defense.
(C-D) The collective estimation task becomes substantially harder in non-uniform environments. In a run with 12 honest robots, final estimates vary widely when tiles are clustered (D) compared to the uniform case (C), illustrating increased variance and higher risk of misclassifying honest robots.}
\label{fig:motivation}
\end{figure}

We draw inspiration from a domain that has already addressed an analogous governance problem outside of robotics: \emph{blockchain-based Decentralized Autonomous Organizations} (DAOs). 
A DAO manages its policies and member interactions through smart contracts, enabling collective decisions and automated enforcing of rules without need for a central authority~\cite{WanDinLi-etal2019:ieeetrans,HasDeF2021:ipr}. 
DAO applications span several fields~\cite{lustenberger2024masteringDAO}: in decentralized finance, DAOs such as MakerDAO and Uniswap govern protocol upgrades, risk parameters, and treasury allocation~\cite{werner2022sok,makerdao,uniswap}; in philanthropy, Big Green DAO and Gitcoin DAO use community voting to support transparent and fair distribution of funds~\cite{biggreendao,gitcoin}; in blockchain infrastructure, DAOs govern protocol parameters, upgrades, and validator incentives~\cite{arbitrum,cosmos}; and in virtual worlds and social platforms such as Decentraland and Yield Guild Games, DAOs let communities collectively shape digital environments and participation rules~\cite{guidi2022decentraland,decentraland,egliston2024cryptogames}. DAOs have also been applied to the collective ownership of real-world assets such as real estate and art~\cite{spychiger2024governance,hunhevicz2021no1s1,lustenberger2025daos}. Across these domains, DAO governance promotes transparency through rules encoded on the blockchain, autonomy via smart contracts that reduce human bias, and member empowerment by enabling participants to influence decisions. Traditionally, DAOs consist of human members; however, it is feasible to design DAOs governed by robots, in which robots acquire and exercise voting power~\cite{DorPacReiStr2024:review}. Such governance could oversee a wide range of swarm activities and serve as the backbone of open swarms, including regulating swarm membership, that is, who may join, remain in, or be removed from the swarm, and enforcing accountability without reliance on a central authority.

\begin{figure}[!ht]
\centerline{\includegraphics[width=1\textwidth]{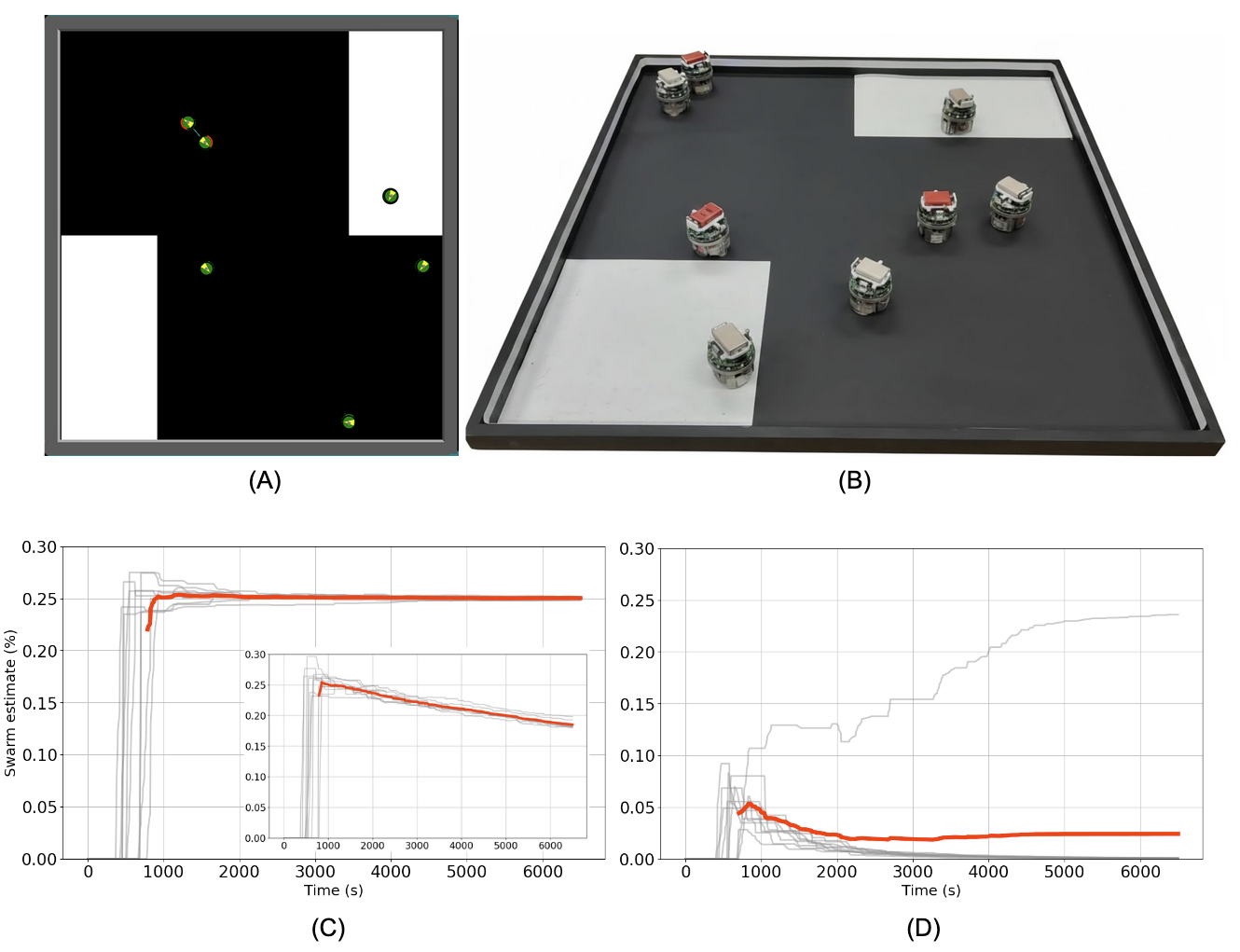}}
\caption{\textbf{Experimental setup and baseline analysis}. (A) The arena, measuring $3.55$m$^2$, used for collective sensing, where Pi-puck robots perform a random walk to estimate the proportion of white tiles (actual value: 25\,\%). Each robot acts as a blockchain node using Toychain~\cite{toychainPacheco2024} to collectively converge on a global estimate. Robots communicate locally within 13~cm. The tiles are clustered in two corners, increasing sensing variance and making the task more challenging compared to a uniformly random tile distribution~\cite{palina2019}. Byzantine robots can further disrupt the task by sending incorrect sensor readings. All experiments are performed in simulation using the ARGoS swarm simulator~\cite{PinTriOGr-etal2012:si}. (B) The physical arena, measuring $1$m$^2$, used for the real-robot demonstrations, with white tiles covering 25\% of the arena.
(C–D) Average swarm estimates in an 8-robot swarm implementing the baseline method of Strobel et al.~\cite{StrPacDor2023:sciencerobotics}, but in an open swarm rather than a closed one. (C) When a naive Byzantine robot is introduced into an initially healthy swarm every 120\,s, the static outlier-based method still converges to the correct estimate. (D) When the base swarm begins with 50\,\% Byzantines and only honest robots are added, the mechanism fails: the on-chain outlier filtering in Strobel et al.~\cite{StrPacDor2023:sciencerobotics} depends heavily on initial estimates of the swarm and accepts only values within 0.10 of the current shared mean estimate. Once Byzantine robots drive the estimate toward 0, newly added honest robots cannot recover the true value. (Inset of C) We also illustrate the failure of the outlier filtering mechanism using `smarter' Byzantines, which send estimates equal to the system mean minus 0.09 to evade outlier detection and shift the estimate towards 0.}
\label{fig:scirob}
\end{figure}

In this work, we present a generic DAO framework for governing participation rights in open robot swarms. 
Through a voting protocol, we provide a proof of concept of how a DAO can be used to form an access-control layer that specifies conditions under which a robot may be removed, specifies which robots are eligible to vote, and determines how votes are processed to execute removal (see section~\nameref{methods}, \nameref{dao}; Figure~\ref{fig:daoimp}). This DAO-based access-control layer can help the swarm achieve two core functions: (i)~identifying and removing Byzantine robots (e.g., due to hardware failure, maintenance needs, or malicious behavior) and (ii)~enabling the swarm to scale in size~\cite{DorTheTri2021:pieee}. We demonstrate this access control layer in a concrete swarm robotics scenario that builds on the collective sensing task introduced by Strobel et al.~\cite{StrPacDor2023:sciencerobotics}, in which a blockchain-enabled swarm seeks consensus on the proportion of white tiles in an environment of black and white regions (see section~\nameref{methods}, \nameref{experimentalsetup}). This study initially demonstrated how blockchains' records can be used for Byzantine detection, however, its outlier detection algorithm assumes a fixed swarm size. Figure~\ref{fig:motivation}A--B shows that, in the collective sensing setup of Strobel et al.~\cite{StrPacDor2023:sciencerobotics}, the outlier detection fails in open swarms, and adapting swarm size as the arena grows improves performance. Moreover, Byzantine outlier detection in Strobel et al.'s~\cite{StrPacDor2023:sciencerobotics} is implemented through transparent on-chain smart contracts and relies on fixed threshold, limiting its resilience in open swarms (see Figure~\ref{fig:scirob}). Robots whose estimates align with the swarm's collective estimate within a threshold are rewarded tokens, whereas inconsistent outliers receive no reward. Since sending estimates requires tokens as a deposit, Byzantine robots are eventually neutralized as their tokens deplete. By contrast, rather than relying on smart contract-based outlier detection to filter Byzantine robots, our approach delegates detection and removal to the DAO. Each voting robot in our access-control protocol independently applies its own decision-making mechanism to select removal candidates. Every robot's detection logic thus becomes a black box to the rest of the swarm, preventing Byzantine robots from inferring or manipulating others' voting strategies or exploiting static thresholds that are publicly visible in smart contracts. 
In our implementation, both honest and Byzantine robots identify candidates for removal using a collusion-detection strategy tailored to the collective sensing scenario; since this detection logic is application-specific, it is not a primary contribution of this work. Newly joined robots must first contribute valid sensing information before gaining
the right to participate in detection and voting, establishing a
probationary phase that bounds the risk posed by new members before they earn trust. We test the practical feasibility of DAO-governed open swarms in simulation, and provide real-robot demonstrations showing that the framework can enable Byzantine-tolerant, scalable, and dynamically growing open swarms.

\section*{Results}

We implement our DAO-based governance mechanism in a blockchain-enabled robot swarm using a simulation framework built on Toychain~\cite{toychainPacheco2024} (see section~\nameref{methods},~\nameref{experimentalsetup}) where each robot is a blockchain node. Toychain implements the same proof-of-authority protocol as  Ethereum’s Clique~\cite{poa:online} and provides similar smart contract functionality, closely replicating the conditions of real-robot experiments previously deployed using Ethereum~\cite{PacStrDor2020:ants, StrPacDor2023:sciencerobotics}. However, since Ethereum is designed for real-time operation, running large sets of simulations with it can be time-consuming. The lightweight Toychain framework overcomes this limitation by synchronizing its clock with the clock of the ARGoS simulator, allowing simulations to run faster than real time and scale to large swarm sizes. Experiments conducted in Toychain simulation enable us to study scenarios that would otherwise be impractical to evaluate with physical platforms due to constraints on time, cost, battery life, and physical space.

Our experimental setup follows the collective sensing task introduced by Strobel et al.~\cite{StrPacDor2023:sciencerobotics}, in which the swarm must estimate the proportion of white floor tiles in the environment, set to 25\,\%. Unlike the uniformly random tile distribution considered in prior work, however, our environment contains clustered regions of white and black tiles (Figure~\ref{fig:scirob}A). As a result, the spatial variance of the robots' local observations is higher (Figure~\ref{fig:motivation}C--D), making consensus more difficult and raising the risk that honest robots are incorrectly identified as Byzantine. 

We examine two types of Byzantine behavior. The first consists of `naive', Byzantine robots that always report a fixed estimate of 0\,\% white tiles. The second consists of `smart' Byzantine robots that attempt to bias the swarm estimate towards zero while attempting to evade detection by exploiting the transparency the smart contracts' outlier detection logic. They achieve this by reporting values that deviate from the swarm mean by only 0.09---slightly below the detection threshold of 0.1 used by Strobel et al.~\cite{StrPacDor2023:sciencerobotics}---while adding small random noise to each vote to avoid sending identical votes, which would be a clear indicator of data tampering or a Byzantine collusion. As shown in Figure~\ref{fig:scirob}C--D, this behavior undermines the outlier-based approach of Strobel et al.~\cite{StrPacDor2023:sciencerobotics}, thereby motivating the need for alternative Byzantine mitigation mechanisms in the swarm.

In our experiments, robots are added to the swarm each time another robot is removed. This condition prevents Sybil attacks, as it limits the rate at which new robots enter the swarm. It is critical that the swarm is able to detect and remove Byzantine members before they are able to overtake the decision process. The swarms ability to do so is controlled by two parameters: $\alpha$ and robot influx.
The parameter $\alpha$ sets the minimum number of valid estimates that newly added robot must contribute before it gains the right to participate in voting. The robot influx is the number of new robots admitted per removal event (see section~\nameref{methods},~\nameref{experimentalsetup}; and Figure~\ref{fig:daoimp}A). In the final 15 minutes of each run, no new Byzantine robots are introduced, so our metric reflects detection and removal alone and does not skew from fresh Byzantines added after each removal. Every 500 timesteps (about 8 minutes), or when prompted by the smart contract, robots independently assess the robots' recent estimates, excluding themselves and the removed robots. The detection mechanism prioritizes suspected collusion among three or more robots submitting identical or near-identical estimates; otherwise, it flags outliers using an adaptive threshold based on the estimates’ standard deviation.
Byzantine robots invert this logic, targeting honest robots whose estimates sit closest to the mean (see section~\nameref{methods},~\nameref{experimentalsetup};).

We evaluate our DAO in three experiments
(see Sections~\nameref{methods} and~\nameref{experimentalsetup},
and Table~\ref{tab:swarm_experiments}):
\begin{description}
  \item[Experiment~1]
    examines whether the DAO can neutralize Byzantine robots
    already present in the swarm;
  \item[Experiment~2]
    examines whether the DAO can stop newly added Byzantine robots
    from overtaking a previously healthy swarm; and
  \item[Experiment~3]
    examines whether the swarm can grow while newly introduced
    robots---which are Byzantine with probability \(50\,\%\)---are removed.
\end{description}

\begin{figure}[!ht]
\centerline{\includegraphics[width=0.75\textwidth, trim={0cm 0cm 0cm 0cm}]{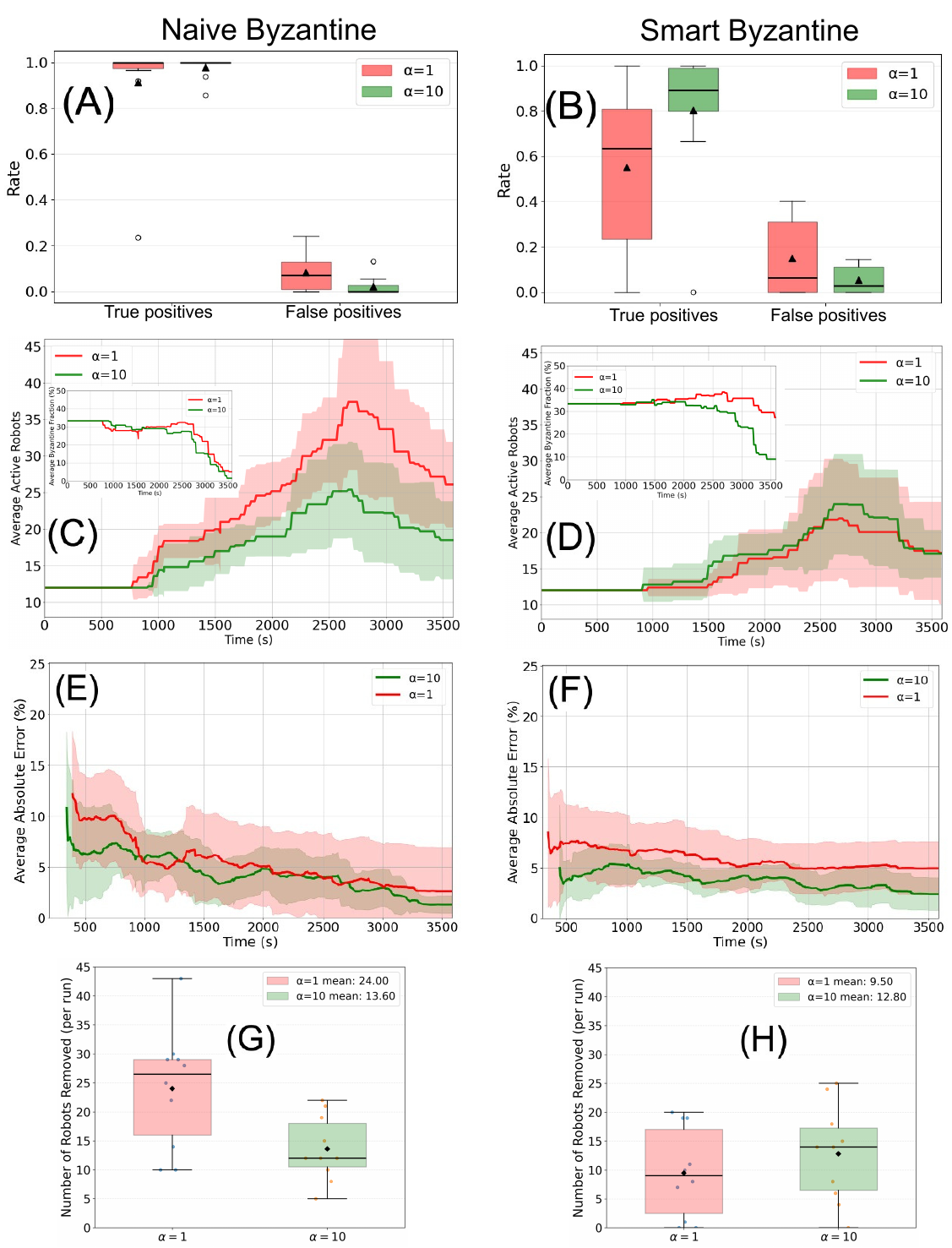}}
\caption{\textbf{Experiment 1 -- Neutralizing Byzantine robots already present in the swarm.} A swarm of 12 robots is evaluated under two types of adversarial behavior: naive Byzantines, which always report an estimate of 0\,\% (A,C,E,G), and smart Byzantines, which report a value approximately 0.09 below the swarm mean (B,D,F, H). For each Byzantine type, we test two DAO settings: $\alpha = 10$ and $\alpha = 1$. The plots show:  (A–B) the classification performance in terms of accuracy (the fraction of Byzantine robots removed from the swarm) and error (the fraction of non-Byzantine robots removed from the swarm); (C–D) the average swarm size over time, with the evolution of the number of Byzantine robots shown in the insets; (E–F) the average absolute error in the swarm's estimates relative to the true value of 25\%, computed at the time when all robots in each run have obtained a mean estimate; and (G–H) the total number of robot removals across the experimental conditions (the black diamond indicates the mean, the horizontal line indicates the median, and the orange or grey dots represent removals in each run). For each value of $\alpha$ and experimental setting (naive and smart), we conduct 10 simulation runs, with the resulting data used to generate all the plots shown in the figure.}
\label{fig:neut}
\end{figure}

\begin{figure}[!ht]
\centerline{\includegraphics[width=0.76\textwidth, trim={0cm 0cm 0cm 0cm}]{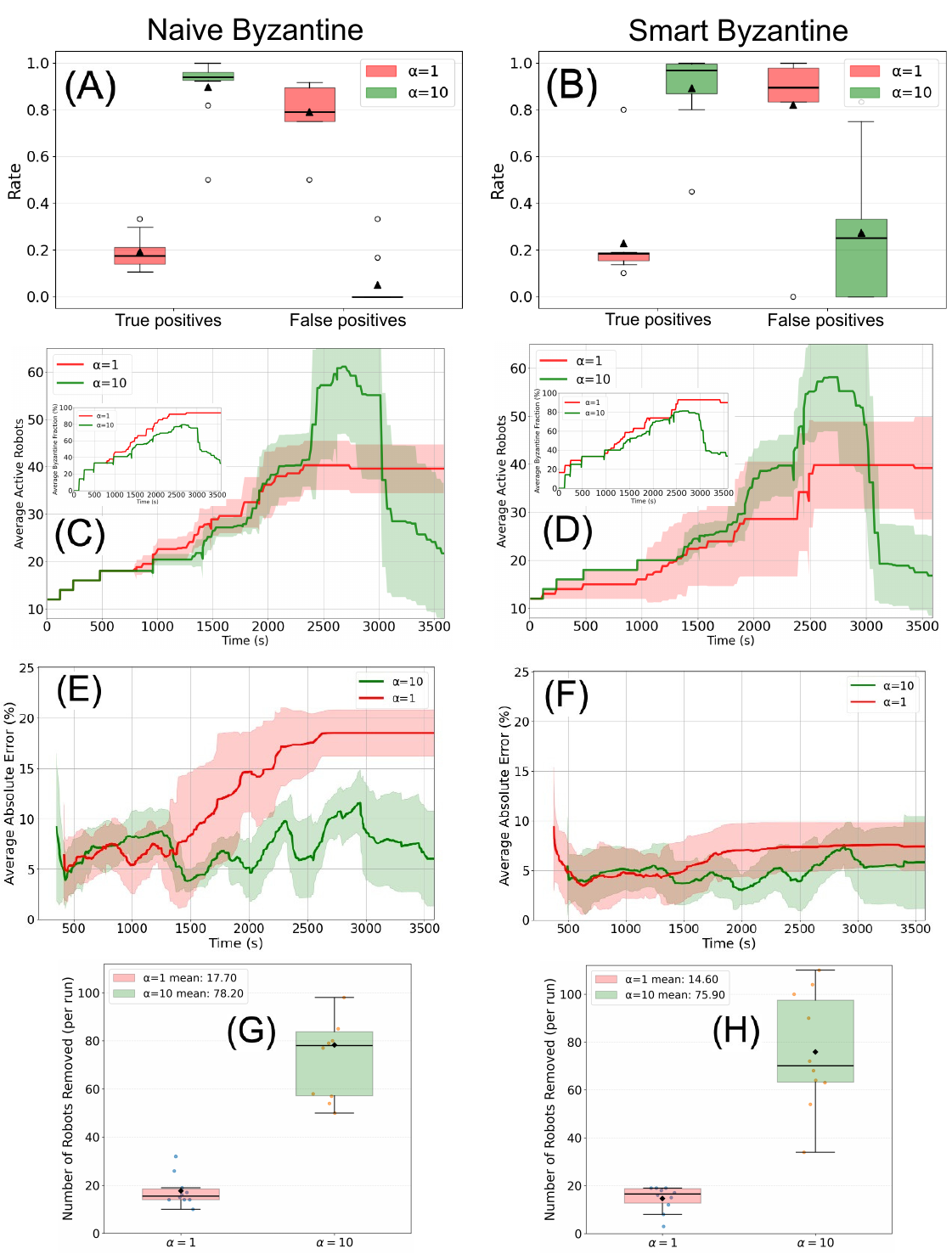}}
\caption{\textbf{Experiment 2 -- Preventing Byzantine takeover.} A swarm of 12 robots is evaluated under two types of adversarial behavior: naive Byzantines, which always report an estimate of 0\,\% (A,C,E,G), and smart Byzantines, which report a value approximately 0.09 below the swarm mean (B,D,F,H). For each Byzantine type, we test two DAO settings: $\alpha = 10$ and $\alpha = 1$. The plots show: (A–B) the classification performance in terms of accuracy (the fraction of Byzantine robots removed from the swarm) and error (the fraction of non-Byzantine robots removed from the swarm); (C–D) the average swarm size over time, with the evolution of the number of Byzantine robots shown in the insets; (E–F) the average absolute error in the swarm's estimates relative to the true value of 25\%, computed at the time when all robots in each run have obtained a mean estimate; and (G–H) the total number of robot removals across the experimental conditions (the black diamond indicates the mean, the horizontal line indicates the median, and the orange or grey dots represent removals in each run). For each value of $\alpha$ and experimental setting (naive and smart), we conduct 10 simulation runs, with the resulting data used to generate all the plots shown in the figure.}
\label{fig:prevent}
\end{figure}

\begin{figure}[!ht]
\centerline{\includegraphics[width=0.75\textwidth,trim={0cm 0cm 0cm 0cm},clip]{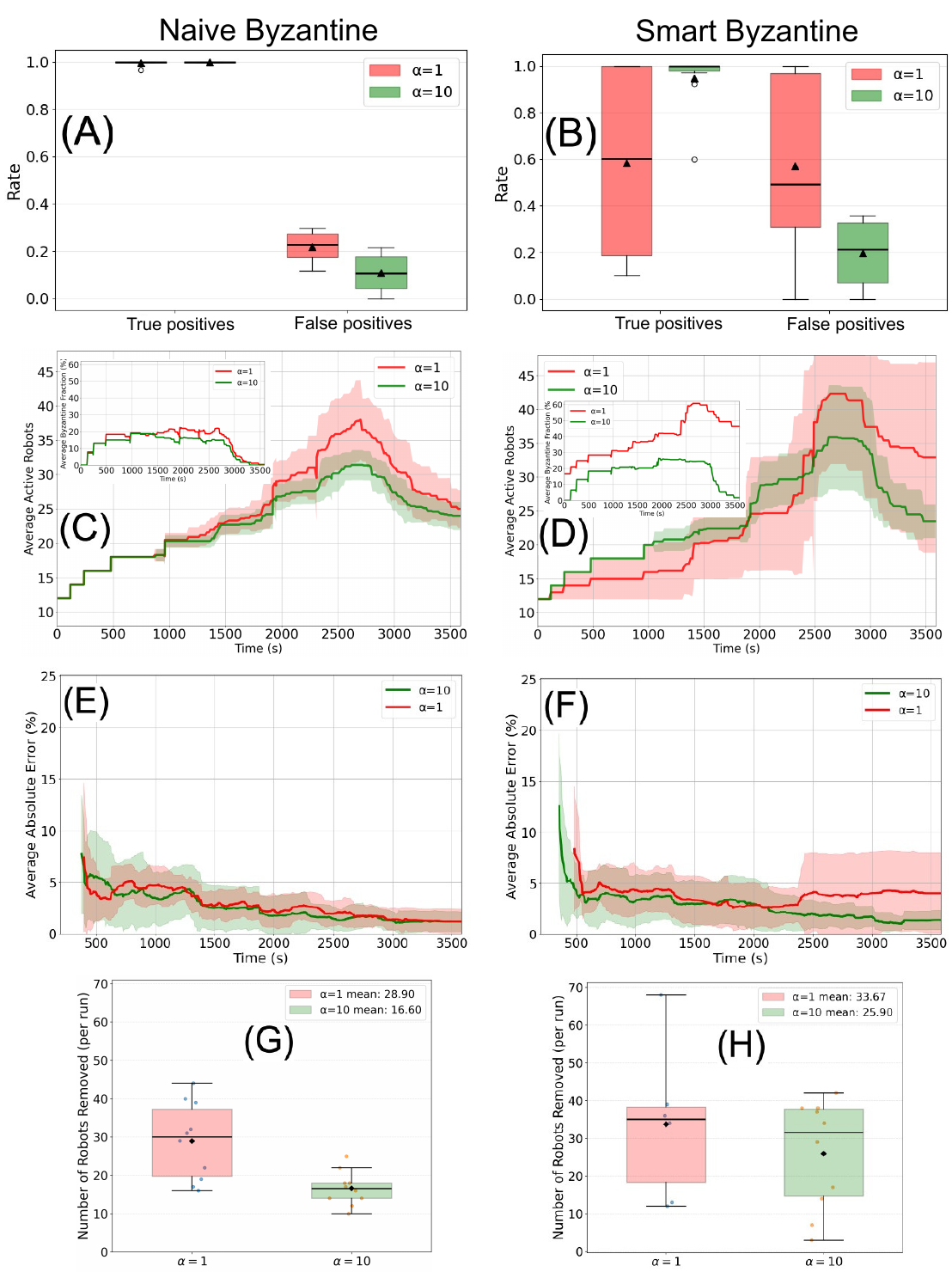}}
\caption{\textbf{Experiment 3 -- Swarm growth under continuous  robot influx.} A swarm of 12 robots is evaluated under two types of adversarial behavior: naive Byzantines, which always report an estimate of 0\,\% (A,C,E,G), and smart Byzantines, which report a value approximately 0.09 below the swarm mean (B,D,F,H). For each Byzantine type, we test two DAO settings: $\alpha = 10$ and $\alpha = 1$. The plots show: (A–B) the classification performance in terms of accuracy (the fraction of Byzantine robots removed from the swarm) and error (the fraction of non-Byzantine robots removed from the swarm); (C–D) the average swarm size over time, with the evolution of the number of Byzantine robots shown in the insets; (E–F) the average absolute error in the swarm's estimates relative to the true value of 25\%, computed at the time when all robots in each run have obtained a mean estimate; and (G–H) the total number of robot removals across the experimental conditions (the black diamond indicates the mean, the horizontal line indicates the median, and the orange or grey dots represent removals in each run). For each value of $\alpha$ and experimental setting (naive and smart), we conduct 10 simulation runs, with the resulting data used to generate all the plots shown in the figure.}
\label{fig:growth}
\end{figure}

Across all three experiments, a clear pattern emerges: robots operating under $\alpha = 10$ consistently achieve a true positive rate (correctly identified Byzantine robots), comparable to or better than those obtained under $\alpha = 1$, while also yielding a lower false positive rate (honest robots misidentified as Byzantines) as shown in Figures~\ref{fig:neut},~\ref{fig:prevent},~\ref{fig:growth}A--B. Because $\alpha$ determines how long newly added robots must wait before participating in voting, a higher $\alpha$ reduces the influence of newly introduced, and potentially Byzantine, robots. This effect is visible throughout Figures~\ref{fig:neut},~\ref{fig:prevent}, and \ref{fig:growth}E--F, where the estimation error is consistently lower for $\alpha$ = 10, most notably in conditions where the goal is to protect an initially healthy swarm from takeover (Figure~\ref{fig:prevent}E--F).

\paragraph{Experiment 1 -- Neutralizing existing Byzantines:}
In the first experiment, the swarm initially contains Byzantine robots, and the system must both remove Byzantine robots and allow the swarm to scale. For naive Byzantines, $\alpha = 10$ yields lower false positives than $\alpha = 1$, while maintaining comparable true positives (Figure~\ref{fig:neut}A). Although both configurations eventually reduce the total number of Byzantine robots (Figure~\ref{fig:neut}C), $\alpha = 1$ yields more false positives, so honest robots are removed alongside Byzantine ones. This is reflected in its higher number of removals per run (Figure~\ref{fig:neut}G) and its larger, more variable estimation error (Figure~\ref{fig:neut}E).  
Smart Byzantines, in contrast, are more difficult to detect because they behave less disruptively than naive Byzantines. The false positive rate is higher for both values of $\alpha$, reflecting their greater subtlety relative to naive Byzantines (Figure~\ref{fig:neut}A--B). Even so, $\alpha = 10$ still improves performance, yielding much higher true positives compared to $\alpha = 1$ (Figure~\ref{fig:neut}B). Despite the lower true positives and higher false positives observed for smart Byzantines, particularly for $\alpha=1$, the Byzantine fraction in the swarm remains around one-third (insets of Figure~\ref{fig:neut}D). Consequently, the swarm's estimation error remains within $\pm5$ percentage points for both $\alpha$ values (Figure~\ref{fig:neut}F), and the number of robots removed per run is similar across the two settings (Figure~\ref{fig:neut}H). 

\paragraph{Experiment 2 -- Preventing Byzantine takeover:}
The second experiment examines whether an initially healthy swarm, that is, a swarm free of Byzantine robots at initialization, can be protected against the rapid introduction of Byzantine robots. We introduce two Byzantine robots at exponentially increasing intervals: after 2, 4, 8, 16, and 32 minutes of the experiment. In addition, each time a robot is removed from the swarm, two more Byzantine robots are added.  
In both the naive Byzantine and smart Byzantine settings, it is evident that $\alpha =10$ yields higher true positives and fewer false positives than $\alpha = 1$ (Figure~\ref{fig:prevent}A--B).
As all newly added robots are Byzantine, the detection task becomes easier for our collusion-based identification method (see ~\nameref{methods},~\nameref{experimentalsetup}). This result contrasts with Experiment~1, in which Byzantine robots were already present at initialization. Here, since the newly added Byzantine robots cannot participate for $\alpha = 10$ removal rounds, the DAO maintains more true positives and fewer false positives. When $\alpha = 1$, however, newly added Byzantine robots can nominate robots to be removed almost immediately, leading to consistently high misidentifications. 
The different classification outcomes are also reflected in swarm size (Figure~\ref{fig:prevent}C--D) and swarm estimation error (Figure~\ref{fig:prevent}E--F). The sharp increase in swarm size and, consequently, in the number of Byzantine robots between timesteps 2300 and 3000 is driven by removals, which trigger an influx of new Byzantine robots. These removals also cause the decrease in the system’s absolute error observed for $\alpha=10$ (Figure~\ref{fig:prevent}E,F). When $\alpha=10$, the swarm does not scale 
to prevent swarm growth driven by an influx of Byzantine robots. This is also shown in Figure~\ref{fig:prevent}G--H, where the number of robots removed per run at $\alpha=10$ is significantly higher. This high removal rate is amplified by the rapid influx of robots, as for every robot removed, two more are added, all of which are Byzantine. The insets of Figure~\ref{fig:prevent}C--D show the resulting population dips followed by sharp increases, driving the Byzantine share of the swarm to as high as 80\% for $\alpha=10$ and 100\% for $\alpha=1$. In the last 15 minutes of the experiment, we can observe that for higher $\alpha$, the fraction of Byzantine robots returns to below 33\,\% (see insets of Figure~\ref{fig:prevent}C--D), whereas for lower $\alpha$ both the Byzantine population and the estimation error remain high. The swarm size (Figure~\ref{fig:prevent}C--D) and estimation error (Figure~\ref{fig:prevent}E--F) eventually appear to plateau for lower $\alpha$ because Byzantine robots can influence removal decisions early, causing the initial set of honest robots to be misclassified and removed. As honest robots are removed, the blockchain protocol may stall~\cite{DorPacReiStr2024:review}, halting the production of new blocks and, consequently, the removal rounds and updates to the swarm mean estimate. This prevents the system from effectively processing further removal decisions. This stall explains why, for $\alpha=1$, the number of robots removed per run remains low (Figure~\ref{fig:prevent}G--H).

\paragraph{Experiment 3 -- Swarm growth under continuous robot influx:}
In the third experiment, we test whether the swarm can withstand the continuous influx of new robots, with each newly added robot having a 50\,\% probability of being Byzantine. As in the second experiment, additional robots are introduced at exponentially increasing intervals (after 2, 4, 8, 16, and 32 minutes).
The resulting true positives are similar to those observed in Experiment 1, where newly added robots also had a 50\% probability of being Byzantine. As before, $\alpha = 1$ performs worse, particularly against smart Byzantines, since newly added robots gain voting power too early, increasing the likelihood that Byzantine robots collectively vote out honest robots. Even for $\alpha = 10$, the false positive rate is relatively higher than in Experiment~1 (Figure~\ref{fig:growth}A--B), once again also reflected in the high number of robots removed per run (Figure~\ref{fig:growth}G--H). 
This is expected, since the influx includes both Byzantine and non-Byzantine robots. As shown in the swarm-size plots, the swarm nearly triples in size for both values of $\alpha$ as new robots, including Byzantine robots, join following removals, before shrinking again (Figure~\ref{fig:growth}C--D). This increased density contributes to greater variance in the estimates, which is already high in the non-uniform environment of our experiments (Figure~\ref{fig:motivation}C--D), inevitably leading to some honest robots being misclassified. The high number of honest robots removed, especially in experiments involving smart Byzantine robots, also stems from our collusion-based detection rule, which flags estimates within a specified threshold of the swarm mean and may therefore flag honest robots as the swarm converges (see~\nameref{method_sec_2}). Despite these false positives, the swarm-size plots also show that the swarm grows from its initial size of 12 while following a controlled downward trend as Byzantine robots are removed. Unlike the more extreme removal observed in the second experiment, swarm growth here remains regulated and balanced, with swarm size ending up almost doubled by the end of the run for higher $\alpha$, where Byzantines are removed and false positives remain low. In particular, when Byzantines are smart, their number decreases to almost zero for $\alpha = 10$ (Figure~\ref{fig:growth}D), whereas more Byzantine robots remain in the swarm for $\alpha = 1$. Thus, although some honest robots are incorrectly removed under the higher $\alpha$ value (Figure~\ref{fig:growth}A--B), Byzantine robots are eventually eliminated effectively and the estimation error remains low (Figure~\ref{fig:growth}E--F). 

\paragraph{Real-robot validation:}

\begin{figure}[!ht]
\centerline{\includegraphics[width=0.95\textwidth,trim={0cm 0cm 0cm 0cm},clip]{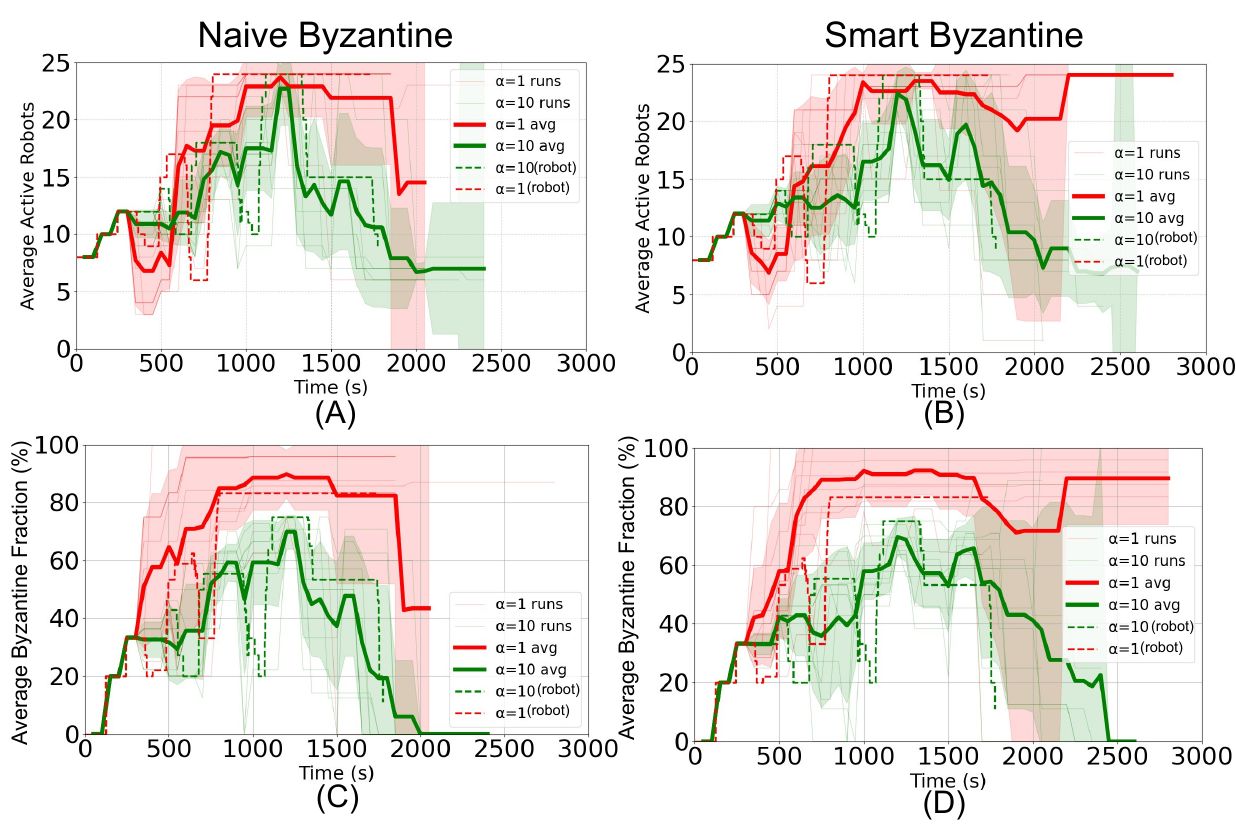}}
\caption{\textbf{Swarm size and Byzantine evolution in simulation and real-robot demonstration} A swarm of eight robots is evaluated under two types of adversarial behavior: naive Byzantines, which always report an estimate of 0\,\% (A–B), and smart Byzantines, which report values approximately 0.09 below the swarm mean (C–D). For each Byzantine type, two DAO settings are tested for $\alpha = 10$ and $\alpha = 1$. The experiments are conducted in a scenario in which a healthy swarm is prevented from being overtaken by newly introduced Byzantines (Experiment 2), and correspond to the real-robot demonstration setup described in Section~\nameref{realrobot}. Each of the four conditions is repeated 10 times in simulation (thick solid lines represent the average; thin solid lines represent individual runs) and the red or green fills represent the 99\% confidence interval of the simulation runs, with results from a real-robot run overlaid for each $\alpha$ value on each plot (dashed lines). The plots show the average swarm size over time (A, B), along with the evolution of the number of Byzantine robots (C, D).}
\label{fig:roboandsim}
\end{figure}

We conducted real-robot demonstrations of the Byzantine takeover prevention experiment, and compared them with simulations configured to match the physical setup. Given the constraints on battery life and the number of available robots, the real-robot and simulation setups were adapted as described in the Methods;~\nameref{realrobot}. Figure~\ref{fig:roboandsim} overlays the real-robot run for all four tested configurations ($\alpha = 1$ and $\alpha = 10$ for both naive and smart Byzantines) on the corresponding simulations. When compared to the insets of Figure~\ref{fig:prevent}C–D, the number of robots decreases in a qualitatively similar fashion across both real-robot setup and simulation runs. The experiment videos are available at \url{https://youtu.be/6EdgD6sOaWM}. 

\section*{Discussion}

From the simulations and selected real-robot demonstrations we conducted, introducing a buffer on voting eligibility, implemented via the parameter $\alpha$---which specifies the minimum number of estimates a robot must contribute before it can participate in swarm-wide voting---favors the removal of Byzantine robots over honest robots more effectively than $\alpha$ = 1 across the three experimental settings. Regulating swarm membership through this rule-based control over who gets to participate in collective decisions, and when, is particularly important because real-world robot swarms cannot always assume that all robots behave as designed. A swarm may include robots from multiple stakeholders with competing objectives, or robots compromised by malicious actors deviating from the expected collective strategy.

The Byzantine detection problem studied here is more challenging than in most prior works, which typically assume a fixed set of Byzantine robots whose identities remain constant throughout the experiment. In all three of our experiments, dynamic swarm membership not only creates more opportunities for Byzantine infiltration but also complicates detection, as the system must continuously reassess trustworthiness under changing conditions. Although real-world scenarios in which a majority of robots are Byzantine may be unlikely, our system can remain operational under such conditions provided Byzantine robots have not yet gained expulsion rights and non-Byzantine robots expel them before they gain control.

In our DAO implementation (Figure~\ref{fig:daoimp}), the smart contract does not determine which robot is Byzantine; instead, it only enforces a collective decision by aggregating votes from eligible robots, each of which can, in principle, identify and accuse peers according to individually-determined criteria. Therefore, the decision of which robots to remove is influenced by the participants rather than solely by the contract's logic. This design contrasts with previous smart-contract-based meta-controllers~\cite{PacStrReiDor2022:ants,StrCasDor2020:frontiers} and aligns it with a key principle of DAO governance: while rules are self-executing, governance remains decentralized and independent of any single authority, including the mediating code itself~\cite{HasDeF2021:ipr}. 

One of the important features of DAOs is to not only operate routine tasks through decentralized participation, but also to evolve by modifying their own rules through democratic processes~\cite{HsiVerAnd2018:jod}. Our implementation, however, decentralizes decisions over a fixed action space: participants can vote on whether to remove or retain a robot, but cannot modify the rules governing this process--the quorum, voting-period length, and eligibility threshold $\alpha$ are determined during the DAO design stage, rather than collectively by the swarm. From the DAO's perspective, this is operational governance without constitutional governance as robots can influence outcomes but not the rules producing them. This distinction is well-recognized in DAO governance literature where on-chain rules can run themselves to execute decisions, but how changes to those rules can be securely governed remains a separate, harder problem~\cite{HsiVerAnd2018:jod,ReiWuiMan2018:topoi}. A stronger DAO implementation would extend towards constitutional governance by allowing robots to vote on and modify on-chain rules, enhancing the capability to adapt to different environments and task requirements. The presence of constitutional governance of this kind opens a new direction for the design of robot swarms, enabling collective rule-making, conflict resolution, and monitoring---three processes identified as central to DAO governance~\cite{VanJan2023:fbloc}---that could be incorporated into robot swarms using the voting mechanisms as the ones outlined in Methods. To provide a holistic overview of DAO applications in robot swarms beyond the proof-of-concept presented in this work, we split our forthcoming discussion around \emph{three} aspects: potential applications, deployment architectures, and accountability and security considerations.

We begin by discussing potential applications of DAOs. Our proof-of-concept demonstrated that a DAO can act as a decentralized access-control layer that enforces the rules governing the admission and removal of robots from the swarm. This shows how DAO-based governance can support swarm-size scaling, while preserving security and resistance to Sybil attacks, in contrast with previous approaches that either overlooked the security problems of open swarms~\cite{higgins2009survey,HunHau2020:nmi}, or relied on secure but inflexible static membership~\cite{StrPacDor2023:sciencerobotics,pacheco2025oracle}. Beyond membership management, a DAO framework has the potential to provide governance mechanisms for a wide range of swarm coordination problems, especially in real-world applications where robots may have competing objectives or act maliciously. We specify below those coordination tasks for which DAOs can be relevant, how using DAOs to address them adds value, and potential challenges of DAOs when applied to them.





(i) Resource allocation: robot swarms often operate in environments where robots must coordinate not only their tasks but also their access to limited shared resources. These may include environmental resources, such as charging stations, storage facilities, or workspace, as well as computational resources, such as communication bandwidth, processing capacity, or shared data services.
Classical foraging and task-allocation models (e.g., response-threshold~\cite{yongming2010swarm,yang2009swarm}) optimize the division of labor among robots, but do not encode rights over these resources or account for disputes between robots competing for resources. A DAO can instead treat scarce shared resources as a common good that the swarm collectively manages. This means giving the swarm explicit, on-chain rules or who can claim a resource, how much it can take, and what happens if it takes too much. These rules follow design principles for managing shared-resource systems, often established in institutional economics~\cite{Ostrom1990:commons} and since been adopted in DAO governance design~\cite{VanJan2023:fbloc}. This could be implemented in market-based resource allocation~\cite{diasmarketrobots2006} in which a robot needing a scarce resource submits a bid, and the highest bid wins because truthful bidding is each robot's best strategy through mechanism design, reflecting genuine need rather than attempts to trick the system. Alternatively, a reputation-based priority mechanism could be used, in which a robot stakes some of its reputation to claim a resource and must first build up that reputation before gaining priority to use it, but loses it if later found to have hoarded the resource rather than genuinely needed it. Current blockchain-based robot swarm solutions can automatically penalize dishonest behavior and reward honest contribution using wealth or reputation~\cite{VanPacStr-etal2023:SciRep, StrPacDor2023:sciencerobotics}. What a DAO adds is the capacity to self-regulate such inter-robot token economies, and resolve disputes through collective judgment, while maintaining on-chain records for posterior analysis. However, detecting misuse of shared resources and designing suitable incentive-aligned protocols is inherently resource-specific. 
Blockchain-based protocols have implemented, for instance, wireless-coverage networks that confirm whether a hotspot is providing genuine radio coverage at a specific location and time by nearby nodes acting as witnesses to validate the beacons~\cite{JagYenWuSchPan2021:helium}, and storage networks which require providers to submit cryptographic proofs that they are continuously and uniquely storing the data they committed to~\cite{Fisch2019:eurocrypt}. A DAO governing physical or communication resources in a robot swarm would likewise need a detection and reward strategy tailored to the resource in question.


(ii) Hierarchies: as robot swarms transition from laboratory settings to real-world deployment, hierarchical organization becomes important for maintaining efficiency and scalability, especially when specialized robots or formation of subgroups are needed under scarce resources or complex environments~\cite{DorTheTri2020:scirobotics}. Existing approaches to hierarchy in robot swarms typically assign roles by hardware specialization~\cite{nitinnatcomm2017}, behavioral differences~\cite{kengyel2015potential} and more recently through dynamic role switching between robots~\cite{zhu2024selforganizing}. A DAO instead can be utilized to govern promotion to leadership roles, using the same propose-vote-execute mechanism we demonstrated. Eligibility for promotion in a hierarchy could require a record of past actions over a specified period (e.g., analogous to our $\alpha$ threshold), a term limit, and a recall vote if the leader underperforms or is later found Byzantine. For instance, this methodology could extend work on mergeable robot swarm morphologies~\cite{nitinnatcomm2017}, in which robots dynamically form and reconfigure self-organized hierarchies with interchangeable `brain' positions that temporarily monitor sensing, actuation, and decision-making for a subswarm~\cite{zhu2024selforganizing}. In that architecture, which robot becomes the brain is decided by quality metrics that can be mission-specific, such as swarm size or coverage area. 
Introducing a DAO governance layer on top would not change the mechanism for occupying the brain position but would make decisions about who occupies it accountable, contestable, and revocable in a way that emerging hierarchies based on local rules do not provide but are required for safe real-world deployments, especially in open or multi-stakeholder scenarios. From a different perspective of DAO design, using DAOs to establish hierarchy in robot swarms may seem counterintuitive, as DAOs are often characterized as inherently non-hierarchical~\cite{HsiVerAnd2018:jod,Davidson_2025}. Recently, however, DAO implementations have emerged that are explicitly hierarchical. Colony, for instance, is a hierarchical DAO framework that organizes members into nested domains with delegated permissions, where authority is tied to reputation earned through contribution~\cite{ColonyDocs}. Hierarchy can also emerge unintentionally: an analysis of deployed DAOs found that early implementations often developed hierarchical structures and power concentrations of power resembling those of traditional firms, despite their decentralized design~\cite{BorMar2024:lit}, highlighting the need for further research into how DAO governance structures can reconcile decentralization with the emergence of hierarchy.

 (iii) Protocol updates: when deployed in the real world, robot swarms will inevitably face dynamic conditions that will require swarm algorithms to adapt as the environment changes~\cite{DorTheTri2020:scirobotics}, such as communication degradations and latency, environmental constraints due to obstacles and scale of area, availability of robots due to computational load, faults and infiltrations. While swarms can adapt their behavior according to local conditions within a fixed algorithm, they generally lack the means to collectively decide on protocol rules and parameters, which are normally set by a human designer before deployment, and to synchronize when robots should enact those changes. 
 A DAO can give the power to enact changes to swarm algorithms that apply globally, and can be accounted for and monitored, enabling constitutional governance. 
This can be potentially achieved by extending the voting pattern we utilized for establishing access control to modulating its own rules, rather than only to swarm membership. Routine removal votes could use a simple majority, as in our current implementation, while high-stakes decisions that change the voting protocol itself, such as amending the quorum size, voting-period length, or eligibility threshold $\alpha$, could require a higher super-majority as they need to be properly vetted. Decentralized-finance (DeFi) governance follows a similar tiered logic, where core protocol and governance changes typically demand higher quorums and longer execution delays than routine parameter adjustments~\cite{aave_documentation,curve_dao_ownership}. A safeguard already used in DeFi is a timelock, a mandatory delay between a passed rule change and its activation, giving the swarm or a human supervisor a window to detect and reject a malicious change before it takes effect~\cite{aave_documentation,curve_dao_ownership}.  Prior work, for instance, has shown that using blockchain in robot swarms can suffer from block production latency, since every transaction must be confirmed and committed to a block before it takes effect. This latency depends on network connectivity and on blockchain parameters, particularly block time, which must be tuned carefully to avoid forks~\cite{SimStrCimDoretal2025:iros}.

(iv) Human-swarm interaction: traditionally, real-time human interaction has been difficult to implement in robot swarms because of their decentralized design. This is also because real-time approaches can end-up operator-centric which risks making the human-in-the-loop a single-point-of-failure or potential bottleneck~\cite{ashcraft_hsi_2019,kolling2016}. Recent literature in swarm robotics, however, highlights the need for ethical governance, as human oversight remains no less important for swarms than it is for single or centralized robotic systems, and this oversight needs to be incorporated in advance of deployment rather than improvised after a failure~\cite{winfield2025ethical}. A DAO-governed swarm can address this by assigning certain decisions to stakeholder committees with transparent roles and clearly defined 
authority, while robots continue to manage routine operational decisions within the policies established through collective voting. This is because in many high-stakes scenarios, robots' algorithmic decision-making may be insufficient for governance decisions particularly those involving ethical trade-offs or significant consequences for stakeholders. This maps onto the operational/constitutional distinction introduced earlier where humans hold the constitutional layer, robots hold the operational layer, and each is accountable to rules recorded transparently on-chain. This creates a new model of human-swarm interaction, where governance evolves through decentralized participation rather than the authority of a single designer or operator, while preserving autonomous operation. Within a DAO both the robots and stakeholder committees will have a bounded governance role. For instance, human stakeholders propose and vote on high-level, constitutional parameters (removal thresholds, quorum sizes, emergency-pause conditions, approved protocol-update procedures, and the categories of action requiring human co-signatures) and passed votes update the smart contract's state accordingly. Robots, in turn, read these on-chain policies voted on by stakeholders, monitor local and peer data, and autonomously initiate operational proposals within the boundaries the humans have set; robot-initiated proposals are then voted on among robots according to the human-defined rules, with outcomes enforced automatically by the contract (e.g., altering robots' voting weight, quarantining it, or removing it). Humans retain the right to intervene as needed to refute, pause, or require manual review, but according to the governance rules they themselves established in advance. DAOs can therefore build a deliberate path for human intervention into the swarms by design, rather than leaving it as an uncontrolled fallback. This intervention is necessary because the same property of automated execution that makes on-chain rules powerful, also means they cannot correct course if the rules themselves turn out to be wrong~\cite{ReiWuiMan2018:topoi}. Here, we extend this same logic to structure the roles of human stakeholders and robots within the governance process itself, rather than treating them in isolation.



(v) Mission control: multi-robot systems typically rely on a central mission controller~\cite{yan2013survey}, while swarm robotics implementations usually operate in a single, predefined behavioral mode (e.g., aggregation, exploration, formation, transport). Complex real-world missions, however, require decentralized
mechanisms to prioritize and switch between such behaviors~\cite{DorTheTri2020:scirobotics,Gupta2024ants,stonge:fri:2020}. Existing approaches to complex missions rely on offline design using evolutionary algorithms~\cite{FraBraBru-etal2014:si,szpirer2024automatic}, and therefore cannot assign mission priorities or switch behaviors in real time. A DAO can govern swarms' mission priorities, as the result of a decision process arising from shared governance, rather than a central operator dictating mission goals, or from the assumption that all robots share a single, predefined objective. Stakeholders or robots, which may have competing or aligned objectives, can propose, vote, and agree on high-level mission goals, 
such as which region to explore, which environmental variables to prioritize, or when to switch between different behavioral modes. Once a proposal is approved, the swarm self-organizes to meet it, switching between tasks or priorities, rather than following a fixed sequential procedure or finite state machine. 
This framework keeps critical high-level decisions securely and transparently enforced through blockchain-based DAO mechanisms, while preserving scalability and autonomy of decentralized swarm control. However, designing mission control involves setting protocol limits, thresholds, and quorums such that mission switching does not create bottlenecks by consuming excessive time, and these parameters must be set accordingly. This is related to the trade-off that shapes distributed decision-making in robot swarms, where speed and accuracy are jointly influenced by quorum size and by how much information is pooled before a decision is made~\cite{ValHamDor2015:aamas,antonic2024collective}. Requiring a vote on every granular decision in DAO governance similarly increases the cognitive and time cost of participation, and therefore, low voter participation is already a widely documented consequence in deployed DAOs~\cite{CongCenGov2024}.

\begin{figure}[!ht]
\centerline{\includegraphics[width=0.8\textwidth,trim={4cm 4cm 1cm 1cm}]{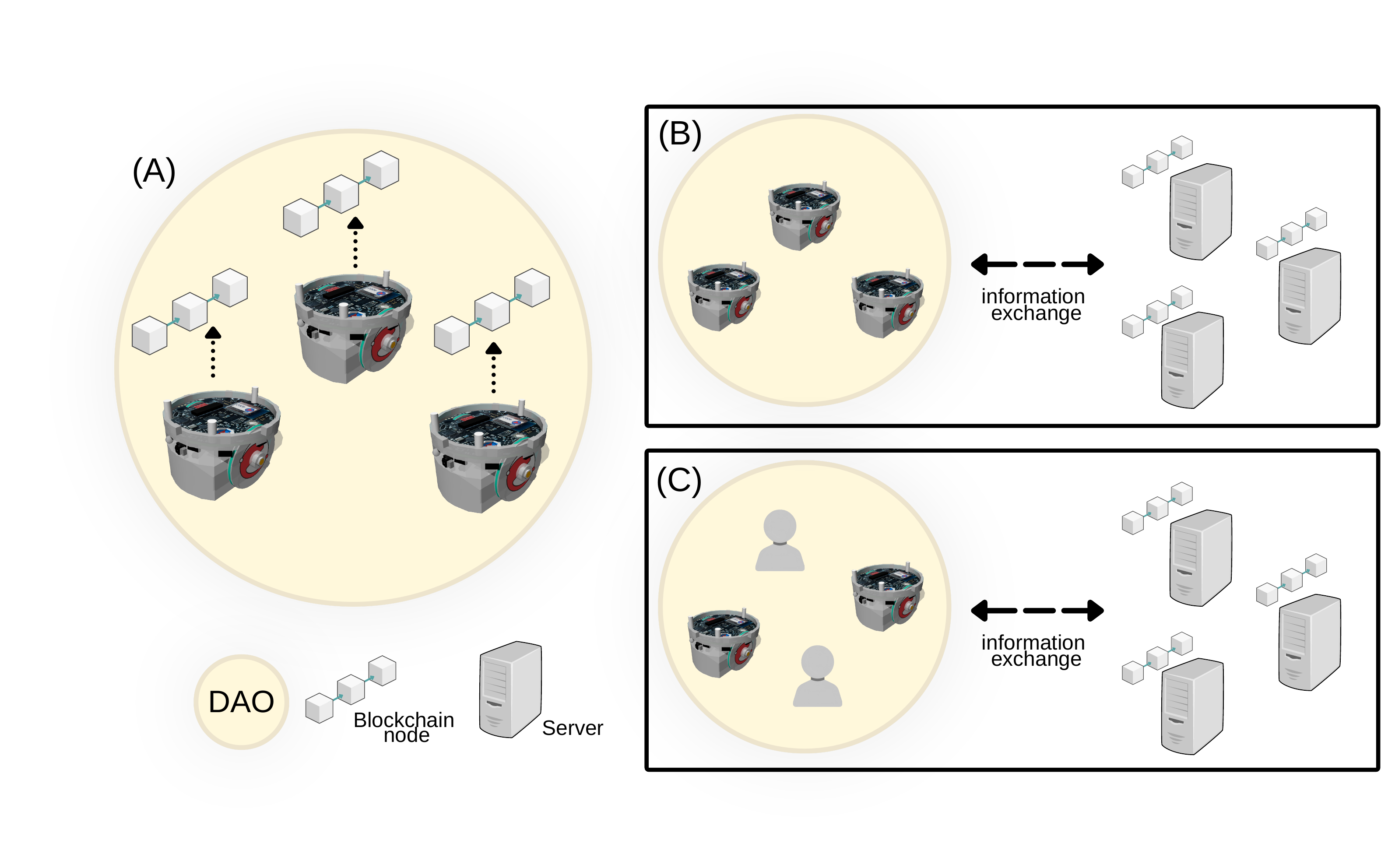}}
\caption{\textbf{DAO architectures.} (A) Swarm-hosted architecture where the robots themselves maintain the blockchain and the swarm is simultaneously the subject of governance and the infrastructure that enforces it. (B) Externally-hosted architecture in which the robots that are part of DAO access external network to participate in governance decisions on a public blockchain, maintained by a distributed set of nodes rather than the swarm. (C) Multi-stakeholder architecture with DAO membership extending beyond robots to include human participants, who take part in the
same governance process alongside the swarm.}
\label{fig:architecture}
\end{figure}

The second important aspect of DAO implementation is its deployment architecture. A DAO's governance rules are enforced through smart contracts and its voting decisions are recorded on a blockchain ledger that provides the technological foundation for decentralized governance; however, where this blockchain infrastructure resides is itself a design choice. In our work, the blockchain hosting the DAO was maintained by the robots themselves (Figure~\ref{fig:architecture}A). As such, the swarm was both the subject of governance and the infrastructure responsible for maintaining it. More generally, however, robots need not maintain the underlying blockchain themselves~\cite{DorPacReiStr2024:review}. If robots have access to an external communication network, they could instead participate in a DAO deployed on a public blockchain, maintained by a much larger and more geographically distributed set of nodes (Figure~\ref{fig:architecture}B). In this model, swarm decisions would be enforced according to the consensus reached by that external network rather than by a single coordinating entity, which eliminates single points of failure, and prevents concentration of control in any individual stakeholder. Extending this further, especially in from a human-swarm application perspective, the DAO's membership need not consist of robots alone: it could include multiple stakeholders, including humans, participating alongside robots in the same governance process (Figure~\ref{fig:architecture}C).

The third aspect of DAO implementation concerns accountability and security. As DAOs leverage blockchain technology, they inherit its properties such as auditability, transparency, and automated execution. The blockchain provides a tamper-proof record of voting actions, and smart contracts ensure that decision rules are executed exactly as specified, reducing ambiguity in decision-making. These properties guarantee that a policy executed by a DAO leaves an immutable trace, especially in implementing constitutional governance, by recording how that policy came to be: which robots proposed it, which voted for it, and under which rule. This attributability is directly relevant to accountability frameworks for swarm robotics, which call for traceable responsibility as a precondition for real-world deployment~\cite{winfield2025ethical}. It also creates a point for ethical review where an auditor can interrogate not only whether a rule was executed correctly, but whether the rule itself was legitimately constituted and whom it privileged.

However, the security of a DAO ultimately depends on the security of its smart contracts~\cite{WanDinLi-etal2019:ieeetrans}, as demonstrated by recent
hacks~\cite{MehShiGia2019:jcit,FeiFriHei2024:aft}. These incidents arise not only from vulnerabilities in the code itself, such as the 2016 exploit of `The DAO,' but also from purely governance-level attacks, including legitimate-looking proposals that conceal malicious code to steal a DAO's data or funds, exploitation of the user-facing interfaces through which humans participate in governance, and spamming the protocol with a high volume of proposals, some of which may themselves be malicious~\cite{FeiFriHei2024:aft}. Implementing DAO governance for robot swarms can carry the same risk. As tasks grow more complex and multifaceted beyond collective sensing, contract complexity increases correspondingly, and this complexity must be carefully managed to avoid introducing similar
vulnerabilities. Our implementation reduces this risk through local, individually-run detection, in which each robot judges incoming estimates on its own rather than relying on information shared across the swarm. Because Byzantine robots cannot see the detection strategies used by their neighbors, this makes it harder for them to coordinate an attack. The same design also helps address delayed Byzantine behavior, in which a robot behaves honestly for a long time to build up trust before turning malicious. Because DAO-based removal mechanisms can judge behavior over a longer time horizon rather than only
recent actions (see~\nameref{methods},~\nameref{dao}), the system weighs a robot's full history, rather than being misled by outdated information or by a few recent but unrepresentative actions.

\section{Methods}\label{methods}
We divide the Methods into two sections. The first section~\nameref{method_sec_1} provides a background on blockchain, smart contracts, and the consensus algorithms, along with the possible mechanism that can be used to form DAOs for access control in a swarm, which can be adapted to other potential applications discussed in the Discussion section. The second section~\nameref{method_sec_2} introduces the simplified DAO we implemented to detect, vote on, and remove Byzantine behavior in a collective sensing scenario, followed by our experimental setup.

\subsection{Blockchain and DAOs}\label{method_sec_1}
\subsubsection{Blockchain fundamentals}\label{blockchain}
Blockchain technology enables peer-to-peer networks to reach consensus on shared data without requiring trust among participating nodes~\cite{Nak2008:techreport,But2014:techreport}. The blockchain is a decentralized ledger composed of sequentially linked blocks, beginning with a common genesis block. Each block records data generated by nodes called \emph{transactions} and includes a cryptographic hash referencing its predecessor. By leveraging blockchain technology, tamper-proof algorithms can be executed in a distributed manner by the use of \emph{smart contracts}. Robot nodes exchange information via blockchain transactions and can establish system-wide rules encoded within these smart contracts, enabling the distributed control and coordination of swarm behavior.

Using the Proof-of-Authority (PoA) consensus protocol~\cite{poa:online}, a designated set of $N$ nodes, known as \textit{sealers}, create new blocks at fixed intervals. After proposing a block, a sealer must wait for at least $\left\lfloor \frac{N}{2} \right\rfloor + 1$ subsequent blocks before producing another block. When multiple sealers generate blocks simultaneously (but potentially in different partitions), conflicting versions of the blockchain may arise due to differences in transactions or their ordering, leading to \emph{forks}. To resolve such conflicts, nodes agree on the chain with the highest difficulty: a difficulty of 2 is assigned to blocks produced by the preferred sealer (for each block, a preferred sealer is selected in a round-robin manner) and a difficulty of 1 to those produced by non-preferred sealers. Any attempt to modify an existing block in a blockchain invalidates all subsequent blocks (and, therefore, reduces the chain's overall difficulty), ensuring data integrity and tamper resistance.

Another feature of PoA is the minimum interval between successive blocks--the block period--which we set to 15\,s by default, in line with prior work~\cite{StrPacDor2023:sciencerobotics,SimStrCimDoretal2025:iros,PacStrDor2020:ants}. In blockchain-enabled robot swarms, however, actual block production often takes longer than this target period, as sparse connectivity and a continually changing network topology can leave no suitable sealer available for extended  periods of time.

\begin{figure}[!ht]
\centerline{\includegraphics[width=0.95\textwidth]{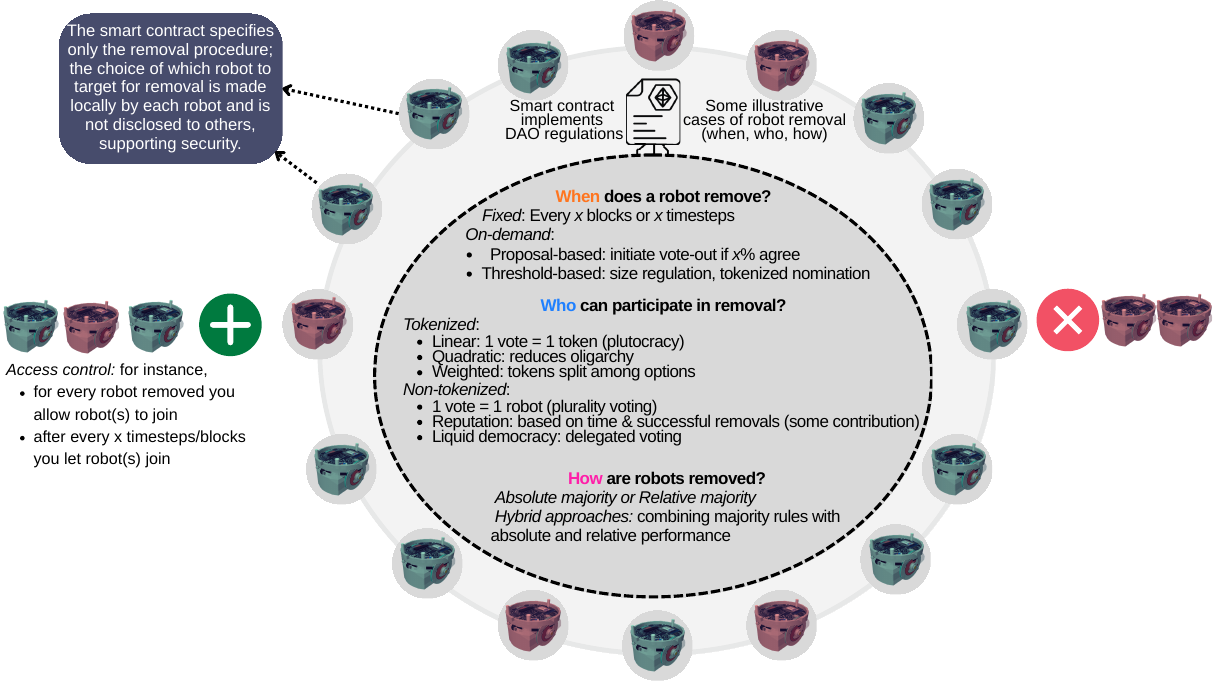}}
\caption{\textbf{DAO-based governance of participation rights in open robot swarms.} Illustration detailing how the DAO regulates swarm membership through voting by defining when removal is triggered, which robots participate in the decision process, and how removal is executed, with a particular focus on Byzantine robots (the e-puck  robot illustration taken from  Webots, Cyberbotics). As underperforming or malicious robots are removed, new robots can be added to maintain or adjust the swarm size. The voting logic used to identify removal candidates can remain private to each robot. Detailed explanation of these mechanisms are provided in section~\nameref{methods}.}
\label{fig:dao}
\end{figure}

\subsubsection{DAO} \label{dao}
The DAO's governance is determined by \emph{when} robots vote, \emph{who} the right to vote, and \emph{how} robot votes are processed. These rules are encoded in the swarm's smart contract, which specifies when a robot can be removed, who is authorized to remove it, and how the removal procedure is executed. Existing DAOs employ a variety of governance and voting mechanisms, including token-based systems, reputation-based models, delegated voting, and hybrid approaches~\cite{ding2023DAOsurvey,yu2019repucoin,tamai2024dao}. These mechanisms can be adapted for use in robot swarms; we provide some examples in Figure~\ref{fig:dao}. Accordingly, we can design our DAOs around three core principles: when, for whom, and how robots vote.

\paragraph{When to remove}~\\
``When to remove'' refers to the conditions under which the DAO's robot members initiate the removal of another robot. In our robot swarm, this typically happens when a robot exhibits malicious behavior, abnormal estimates, or actions inconsistent with the swarm's objectives. The methodology used by robots to detect malicious behavior can differ from robot to robot, functioning as a black box that prevents others---including Byzantine robots---from learning the detection strategy and circumventing it. In our implementation (section~\nameref{experimentalsetup}), the non-Byzantine robots use a mechanism that detects collusion among robots whose estimates fall within a specific range, enabling them to flag and remove malicious robots. Alternative Byzantine detection methods could range from simple thresholding to filtering out robots whose estimates fall within suspicious ranges, to more complex machine-learning–based approaches that optimize the detection of malicious behavior (given the wide range of alternatives possible, developing such detection methods is outside the scope of this paper). The Byzantine robots, conversely, attempt to behave in a way that counters this adaptive thresholding strategy. 

The ``when to remove'' mechanism need not be limited to malicious behavior; it can also be used to regulate the size and efficiency of the swarm. In situations where resources must be conserved or where blockchain performance, such as block latency~\cite{zak_ants26}, or block inconsistency~\cite{SimStrCimDoretal2025:iros}, no longer improves with additional robots, the swarm can be triggered to remove certain robots from the swarm, allowing the system to adapt dynamically. We elaborate on three primary mechanisms for triggering a removal process below. 

\emph{Scheduled removal:}
The robots are allowed to vote on removals at predefined intervals, such as every $x$ timesteps or after $k$ blockchain blocks, during which the robots in the swarm inspect task-specific data received from the robots for signs of Byzantine behavior. This strategy is useful in systems susceptible to a high number of Byzantine robots or abnormal estimates, and depending on the influx of Byzantine behavior, the interval or blocks need to be tuned accordingly.

\emph{Proposal-based removal:}
A removal process is triggered when robots detect Byzantine activity and collectively propose initiating a vote-out round. This mechanism is inspired by quorum-based voting~\cite{karjalainen2020quorum}, in which the removal proposal is activated only if a predefined number of robots (the quorum threshold) signal the presence of a Byzantine robot. While proposal-based voting is more decentralized than scheduled removal and encourages active participation by spreading the power to initiate a vote-out across members, it introduces several challenges. Setting the quorum threshold is particularly sensitive, as if the quorum is too high, many valid proposals may fail due to insufficient participation. If the quorum is too low, Byzantine robots could repeatedly signal removal and collude to remove honest robots. 

\emph{Threshold-based removal:}
A removal process may also be initiated through threshold-based mechanisms, in which the process is triggered only after a predefined condition is satisfied. One example is size-based regulation, where reaching a threshold related to the swarm’s ideal size can initiate a vote-out process to prevent the swarm from becoming excessively large. This threshold may be defined with respect to the area that needs to be covered, physical congestion, communication bottlenecks, operational cost, or stricter security requirements that arise from avoiding redundant units. 

\paragraph{Who removes}~\\
``Who can request or approve removal'' defines how authorized robots within the DAO initiate or participate in removing a member from the swarm. These authority rights can be structured using either tokenized or non-tokenized governance mechanisms. In tokenized systems, governance tokens form the basis of decision-making where robots that hold these tokens can vote on proposals, influence collective actions, and shape system-level decisions such as software upgrades, parameter changes, or resource allocations. 
In non-tokenized approaches, voting power is not linked to token ownership. Instead, a robot's authority to propose or approve removal is derived from performance-based metrics such as historical sensing accuracy, behavioral reliability, or sustained participation in swarm tasks.
In the following, tokenized and non-tokenized mechanisms are described in more detail.

\emph{Tokenized mechanisms:} In tokenized mechanisms, voting power grows with the number of tokens a robot holds. Applied to robot swarms, such tokens can be distributed initially as a form of universal basic income, as shown in Strobel et al.~\cite{StrPacDor2023:sciencerobotics} and other blockchain-based swarm robotics studies~\cite{VanPacStr-etal2023:SciRep,pacheco2025oracle}, and later earned through useful work or consistently honest behavior. The simplest form of tokenized mechanism is the familiar \textbf{`one token, one vote'} approach, where greater token ownership directly increases influence. Although this design can motivate participation, it also introduces the risk of centralization~\cite{kaal2021blockchain}. If a few robots accumulate most of the tokens, the system can drift toward a plutocratic structure, which becomes particularly problematic when those high-token robots behave maliciously. To counter this, \textbf{quadratic voting}~\cite{lalley2018quadratic} offers a more balanced alternative. Quadratic voting increases the cost of casting extra votes quadratically, resulting in large token holders retaining influence but not easily overpowering token holders with fewer tokens. This encourages honest robots to use their tokens carefully and greatly reduces the risk that a small group can dominate the voting process.
When a swarm needs to choose among several options, such as identifying many robots in the swarm that might be behaving in a malicious manner, \textbf{weighted voting} can be a flexible voting mechanism. In weighted voting, each robot can distribute its tokens among different options, reflecting varying levels of confidence rather than a single binary choice. The effect of weighted voting depends on how each robot distributes its available tokens among the potential candidates. 
As robots must also reserve tokens for essential tasks, such as contributing accurate environmental estimates, newly arrived robots have limited freedom in how aggressively they can vote. This naturally restricts their influence and prevents them from immediately coordinating a strong attack. As a result, Byzantine entrants cannot rapidly sway the collective judgment or overwhelm the system with malicious nominations. 
Moreover, since robots earn tokens by voting for the same candidates as the majority and by successfully removing faulty robots, the voting process reinforces honest behavior and makes it costly for newcomers to act adversarially.


\emph{Non-tokenized mechanisms:} In this case, the DAOs use governance models where the voting power of robots is not tied to token ownership. Several approaches from voting literature can be utilized for allocating voting rights in such swarms. The simplest model is \textbf{plurality voting}~\cite{sharma2024unpacking,ding2023DAOsurvey}, where each robot is assigned exactly one voting right. Although easy to implement and intuitively fair, plurality voting is vulnerable to manipulation and may produce outcomes that do not reflect the swarm's true preferences. For example, depending on the quorum, a removal can occur with only a small fraction of overall support, stemming from Byzantine or malfunctioning robots. In a permissionless swarm, plurality voting introduces an additional challenge: any robot can create new identities, making the system vulnerable to Sybil attacks. An adversary could simply generate multiple identities to amplify its influence and skew the outcome of collective decisions.

More sophisticated non-tokenized systems assign voting power based on reputation, contribution, or time spent doing a task rather than token accumulation. In \textbf{reputation-based voting}~\cite{kaal2021blockchain,yu2019repucoin}, robots can build reputation by providing accurate estimates, contributing useful work (including physical work using oracles~\cite{ZhaPacStr-etal2023:iros}), or remaining in the system without being voted out. Robots with long-standing, consistent behavior earn more influence, making the swarm more resistant to newly introduced or low-reputation Byzantine robots. Reputation can also be set to decay over time to prevent early contributors from permanently dominating governance, to ensure opportunities for newer robots that may not be Byzantine, and to keep the swarm adaptive to environmental changes or shifts in robot capabilities. By doing so, Byzantine robots cannot exploit long-term trust by behaving well initially and becoming malicious later. 

Another non-tokenized mechanism is \emph{delegated voting}, in which robots that lack confidence in their local estimates transfer their voting power to other robots with stronger sensing capabilities or higher demonstrated accuracy. Since Byzantine robots progressively lose reputation or rewards as their behavior diverges from the swarm's truth, they are unlikely to attract delegated votes, shifting trust naturally toward robots with consistent and reliable performance. \textbf{Liquid democracy}~\cite{fan2024dao} implements this idea by allowing newly added or uncertain robots to delegate their voting rights to more experienced peers. In turn, they receive a share of the reward when their chosen delegate successfully identifies and removes a Byzantine robot, creating an incentive structure that aligns individual choices with accurate governance. Such delegation mechanisms also provide a pathway for newly joined robots to gradually accumulate influence (or tokens in tokenized mechanisms), enabling them to participate more meaningfully in future voting rounds.

\textbf{Multisignature voting}~\cite{jungnickel2026dao,ding2023DAOsurvey} can also be adapted to the operational structure of robot swarms. In this approach, a predefined subset of robots, such as the initial swarm members or trusted sealer nodes, must jointly approve a removal decision or validate the votes a robot submits before they are recorded on-chain. This strategy provides strong control over swarm membership, particularly when the initial group of signers is verified to be non-Byzantine at deployment time. Requiring multiple signatures increases accountability and helps isolate malicious behavior, as discrepancies in signature patterns can reveal which robots are acting dishonestly. Nevertheless, multisignature schemes introduce the risk of collusion among the designated signers. A coordinated subgroup could authorize harmful removals or obstruct legitimate ones, which becomes especially problematic in open swarms where robots may be owned or operated by different stakeholders~\cite{StrCasDor2020:frontiers,DorPacReiStr2024:review}.

Finally, \textbf{conviction voting}~\cite{ding2023DAOsurvey} also offers an alternative mechanism suited to robot swarms. In this approach, voting power increases gradually the longer a robot maintains the same vote on a suspected Byzantine robot. Once a voting period is opened by a smart contract, any robot in the swarm can cast or update its vote at any time. A vote that remains stable accumulates `conviction' asymptotically, where at each time step it increases by a fixed fraction of the remaining distance to its maximum value (set by the vote’s weight). If a robot changes its vote, the conviction associated with its previous choice begins to decay. Therefore, the conviction process captures persistent agreement within the swarm while suppressing sudden or coordinated surges that might arise if several malicious robots join simultaneously. By rewarding consistent, time-sustained identification, conviction voting helps the swarm converge on reliable identification of malicious robots. 

\emph{Use cases of tokenized and non-tokenized mechanisms:} The best voting mechanism for a DAO in robot swarms strongly depends on the intended application. Voting is not limited to identifying or removing Byzantine robots but can support other tasks, such as dynamic task-allocation based on robot capabilities, budgets, or coordinating environmental coverage. For instance, robots may bid or stake to accept jobs that match their sensors, battery levels, or functional modules, while smart contracts handle task assignment and verify successful completion. In such scenarios, token-weighted or reputation-weighted voting can ensure that the most capable and reliable robots have proportionally greater influence on which robots are selected for each task. These voting mechanisms also allow DAOs to form open swarms, enabling newly arrived robots with relevant sensors or upgraded hardware to join and gradually gain influence through token accumulation or contribution-based reputation, as in conviction-style voting. Similarly, swarm-level parameters such as environmental exploration rates, communication ranges and sensing thresholds, or repulsion and attraction thresholds can be adapted during runtime through DAO proposals. For these updates, conviction or multisignature voting can help the swarm converge on stable, consistent parameter choices over time, while non-tokenized voting provides a faster alternative when faster consensus is needed.

Hybrid voting mechanisms could combine token and reputation systems, where each system could serve different goals depending on the DAO's purpose, size, and community dynamics. For example, tokens could represent economic commitment, and reputation could capture a robot's historical accuracy, task performance, and consistency in the task at hand. By combining the two through reputation-weighted token voting, a robot’s token-based voting power is scaled by its demonstrated reliability, preventing newly added robots from exerting disproportionate influence before they have proven themselves. Dual-quorum designs could extend this idea by requiring that proposals satisfy both a token threshold and a reputation threshold, ensuring that decisions are supported not only by robots with many tokens but also by robots with demonstrated work. Conviction-based tokenized staking can be another hybrid variant in which robots stake tokens on their preferred outcome, and their stake is amplified by reputation multipliers derived from past task success or diagnostic correctness. Overall, these hybrid approaches can balance the contributions of long-standing, trustworthy robots with the openness needed for new robots to join and eventually participate meaningfully, ensuring that both commitment and demonstrated reliability shape collective decisions. Such hybrid schemes also illustrate that voting mechanisms can be adapted to the specific task and performance requirements at hand, and that possible designs are non-exhaustive.


\paragraph{How to remove}~\\
There are certain mechanisms by which a DAO can enforce a decision based on the votes it receives from the robots (those `who' can remove). Once the DAO has collected the votes, it must aggregate them and determine which robot(s) should be removed from the swarm. A straightforward approach is to apply a simple majority rule, where the robot receiving the highest number of removal votes is excluded from the swarm. This can be refined by introducing threshold-based criteria, such as an absolute threshold (a predefined number of votes required for removal) or a relative threshold scaled to swarm size (e.g., a robot must receive votes from a quorum of swarm members to be removed). For voting mechanisms that produce weighted outcomes, such as token-based or reputation-based voting, the smart contract can instead remove robots whose aggregated score exceeds a predefined removal boundary, ensuring that both the intensity and distribution of suspicion are taken into account. For instance, when robots nominate multiple candidates, the system can instead remove any robot that appears in at least $L$\,\% of all submitted lists, ensuring that widely suspected nodes are prioritized. The decision rules may consider only votes from the current round, or they may incorporate historical information by taking into account each robot's past voting results, allowing the DAO to balance short-term consensus with longer-term behavioral trends when determining removals. For instance, instead of removing a robot immediately, the DAO could run several voting rounds spaced in time. A robot would then be removed only if it is consistently flagged across multiple rounds.

Once removal is decided, the smart contract marks the robot as removed and broadcasts this state to the swarm, which then moves the robot out of the swarm and ceases its blockchain node. Robots update their list of active robots and cease interacting with the removed robot and disregard its information, effectively excluding it from collective swarm behavior.

\subsection{Experimental setup}\label{method_sec_2}
\subsubsection{DAO implementation}\label{experimentalsetup}
In our DAO implementation (Figure~\ref{fig:daoimp}B), which is built on top of the collective sensing setup, the robots evaluate the swarm for Byzantine behavior every 500 timesteps (approximately every 8 minutes) to signal the contract or when they receive a signal from the smart contract to vote on potential Byzantine peers. In our setup, every robot uses the same detection mechanism. In practice, however, each robot could use any mechanism without disclosing it to its peers, because each robot independently analyzes the recent estimation data used in the swarm aggregation process to compute the swarm estimate. The analysis of recent estimates is restricted to the most recent voting rounds (i.e., the estimates robots send that are used to compute the system estimate by averaging) to emphasize current behavior and reduce the influence of outdated estimates. Robots that have already been removed and the local robot itself are excluded from evaluation. The detection logic prioritizes collusion as the strongest indicator of Byzantine behavior. If three or more robots submit identical or near-identical estimates---matching exactly or within a small system-defined tolerance of each other (the deviation threshold use to detect collusion in the system is $\pm0.07$ from the swarm mean)---they are flagged. Such agreement is unlikely among honest decentralized agents and suggests coordinated malicious behavior. If no collusion is detected, the algorithm applies an adaptive statistical filter. It computes the spread of recent estimates using the standard deviation and derives a dynamic threshold scaled by a tunable parameter. Robots whose estimates deviate from the system mean beyond this threshold are considered outliers and selected for removal. This adaptive mechanism prevents honest robots from being penalized during periods of high natural variability but also excludes robots that are sending estimates that are very off. While honest robots flag peers whose estimates diverge excessively from the mean, Byzantine robots do the opposite and always use adaptive thresholding to check for estimates that are closer to the mean, in an attempt to remove honest robots from the system. Overall, the removal mechanism is a simple approach specific to our system that identifies irregular patterns of coordination or deviation, but it is not the key contribution of this work.

Our DAO-based access control implementation has three phases: propose, vote, and execute. 

\textit{Propose} If a robot is detected to be acting maliciously, the honest robots signal the smart contract by executing the \texttt{propose()} function, proposing that a robot needs to be removed. Once a robot in the swarm proposes a vote-out round, the smart contract waits until $x$\,\% (in our case 50\,\%) of the swarm has proposed removal for a robot before proceeding. Robots are eligible to participate in this voting process only if they were present at system initialization (the base swarm) or, if added later, have contributed to the task at least \emph{$\alpha$} times (i.e., submitted $\alpha$ valid estimates). In DAO-based organizational implementations, statutory quorums are often incorporated into the proposal process to ensure robustness, which means that a proposal must not only originate from at least \emph{x}\,\% of the robots but also surpass a predefined minimum threshold of supporting proposals. In our system, this threshold is set to five proposals. Setting the threshold too high can make the Byzantine removal decision-making process too restrictive and time-consuming before the robots are removed. On the other hand, setting the threshold too low can result in frequent voting triggers that could flag non-Byzantines to be removed.

\textit{Vote} Once the smart contract issues a request for votes, each eligible robot compiles a list of robots it considers Byzantine and executes the \texttt{sendvote()} function of the smart contract to participate in the vote-out. Robots that initially proposed a voting round also recompile their removal lists when voting begins to account for new estimates received while awaiting quorum and the smart contract’s initiation of voting. The smart contract waits for five blocks after initiating a vote-out, called voting period. Only robots with the required participation history can vote to remove another robot.

\textit{Execute} After a voting period of five blocks to collect the removal lists, the smart contract removes any robot that appears in at least 50\,\% of the submitted lists and receives at least five votes.

Given that the task at hand is the collective sensing scenario implemented in Strobel et al.~\cite{StrPacDor2023:sciencerobotics}, the smart contract has additional functions that aggregate the swarm estimates of the environment to produce the collective estimate (see Supplementary material). Robots send an estimate of the proportion of white tiles they believe exist in the environment by executing a smart contract function. Each robot estimate is stored in a temporary list maintained by the smart contract. Once this list contains a specified number of estimates, the estimate collection round ends, and the smart contract computes their arithmetic mean, which becomes the swarm estimate before the next round of estimate collection begins. Not all robots vote in every round, as sending an estimate requires staking tokens. When the swarm updates its mean after a removal round, it excludes the estimates of the removed robots and does not take them into account in the subsequent means. After a robot is removed, its tokens are redistributed equally among the robots who contributed to voting it out.
\subsubsection{Platform} 

Our experiments are carried out in the ARGoS swarm robotics simulator using Pi-puck robots. The arena is a square environment measuring 3.55 m$^2$ and enclosed by walls (see Figure~\ref{fig:scirob}A). Unless otherwise specified, each simulation runs for 3,600 timesteps (with 1 timestep corresponding to 1~s, so each run lasts an hour), and we conduct 10 repetitions for each experimental condition. During the final 15 minutes of each run, no additional Byzantine robots are introduced, ensuring that the observed accuracy and misclassification rates reflect the system’s ability to detect and remove Byzantines without being skewed by new Byzantines added immediately after each removal.

Every robot in the swarm operates as a blockchain node, while only designated sealer robots produce blocks using Toychain~\cite{toychainPacheco2024}---a lightweight blockchain framework designed specifically for swarm robotics research. 
The code used in this study is open-source and provided as Supplementary Material.
At the beginning of each run, robots are randomly placed in the arena (see Figure~\ref{fig:scirob}A--B) and execute a random-walk behavior comprising periods of straight motion with random turns,  consistent with prior blockchain-enabled swarm robotics work~\cite{StrPacDor2023:sciencerobotics,StrCasDor2020:frontiers,StrCasDor2018:aamas,SimStrCimDoretal2025:iros}. Only the robots that the swarm is initialized with at the beginning of the experiment are sealers. Any other robots added during the runtime are non-sealers, i.e., they do not produce blocks. Blockchain synchronization among robots occurs locally, within a communication range of 13~cm.

\begin{table}[htbp]
    \centering
    \caption{Overview of the experimental parameters across the experimental cases.}
    \label{tab:swarm_experiments}
\begin{tabular}{|l|p{3cm}|p{3cm}|p{3cm}|    }   
        \hline
        \textbf{Parameter} & \textbf{Experiment 1} & \textbf{Experiment 2} & \textbf{Experiment 3} \\
        \hline
        Initial swarm size & 12 & 12 & 12 \\
        \hline
        Initial number of Byzantine robots & 4& 0& 0 \\
        \hline
        Robot addition rate (without removals)&   0&Exponentially increasing intervals (2 min, 4 min, 8 min, 16 min, 32 min) &  Exponentially increasing intervals (2 min, 4 min, 8 min, 16 min, 32 min)\\
        \hline
        Robots added per removal event & 3&2 & 2\\
        \hline
        Probability of an added robot being Byzantine & 50\% & 100\% & 50\% \\
        \hline
        Alpha values tested &$\{ 1,\,10\}$ &$\{ 1,\,10\}$ &$\{ 1,\,10\}$ \\
        \hline
        Evaluation interval for removal (timesteps) & 500 &  500 & 500 \\
        \hline
        Total experiment duration (timesteps) & 3600 & 3600&  3600\\
        \hline
    \end{tabular}
\end{table}

\subsubsection{Experimental cases}
Using the DAO, we conduct three experiments, which we discuss below. Table~\ref{tab:swarm_experiments} provides a general overview of the differences between them.

\paragraph{Neutralizing existing Byzantines}
We begin with a swarm of 12 robots, of which 33\,\% (4 robots) are Byzantines. For every robot removed, we add three new robots to the system that have a 50\,\% probability of being Byzantine (therefore, the swarm size increases). Newly added robots, whether Byzantine or non-Byzantine, must satisfy the DAO's contribution rules before being granted voting privileges to remove another robot. Using this scenario, we show that a DAO can be used to remove Byzantines that have been part of the swarm since the beginning, as well as new ones that may join the swarm at a later time.

\paragraph{Preventing Byzantines from taking over a functioning system} We start with a swarm initially containing only non-Byzantine robots, but add exclusively Byzantine robots. As Byzantines are added at exponentially increasing intervals---after 2, 4, 8, 16, and 32 minutes---the system detects and removes them to restore convergence on the correct swarm estimate. 
Unlike the first experiment, where three robots (not necessarily Byzantine) were introduced per removal, here only two robots are added per removal, all of which are Byzantine. This compensates for the additional influx from robots added at exponentially increasing intervals, preventing an uncontrolled accumulation of Byzantine nodes. 

\paragraph{Supporting controlled system growth}
We start with an initially healthy swarm and introduce new robots at exponentially increasing intervals, each with a 50\,\% probability of being Byzantine. For each robot removed, two new robots are added, allowing the swarm to grow while maintaining resilience.

In general, besides the periodic check for Byzantine behavior in the system, other parameters also influence the rate at which robots are added and removed from the system to prevent Byzantine takeover. 
The $\alpha$ parameter determines how long newly added robots must remain in the system and contribute valid estimates before they gain the privilege to participate in removals. In our experiments, we test a low value of $\alpha=1$ where a robot needs to contribute only a single estimate to participate in the removal of other robot(s), and a higher value of $\alpha=10$ where a robot must contribute ten estimates without being removed from the swarm before it attains this removal privilege. The influx of new robots in the swarm is also regulated by controlling the number of robots admitted per removal and, in Experiments 2 and 3, by introducing robots into the initially healthy swarm at exponentially increasing time intervals. If most newly added robots are expected to be Byzantine, the inclusion of new robots should be low; if the probability of additions is small, it can be higher. The influx of robots into the swarm is also partially impacted by the interval at which the system removes a Byzantine, which triggers the addition of new robots in the system. 

\subsection{Real-robot demonstrations} \label{realrobot}

We conducted real-robot experiments for the second experimental case, the most challenging scenario, in which a Byzantine takeover must be prevented within an initially healthy swarm. As in real-world deployments, the experiments were subject to practical constraints, including limited robot battery life, a limited number of available robots, and the time overhead associated with removing and replacing robots within the swarm. The experimental setup, described in Section~\nameref{experimentalsetup}, was therefore adapted to account for these physical limitations. Specifically, the following key modifications were introduced: (i) to accelerate the experiment, robots detect Byzantine behavior every 150 time-steps (every two and half minutes), are provided with an additional token alongside the basic tokens provided at the start of each run
; (ii) following a robot removal, a two-minute buffer is enforced before replacement robots are introduced, allowing sufficient time to complete all removals within a removal round; (iii) due to the limited number of robots available, no additional robots are introduced once 24 are active in the swarm, after which only removals take place, either for two rounds or for a maximum of 15 minutes; (iv) for the same reason, the initial swarm consists of eight robots operating within a smaller arena of $1\,\mathrm{m}^2$. One demonstration was conducted for each of four configurations: naive Byzantine with $\alpha = 1$ and $\alpha = 10$, and smart Byzantine with $\alpha = 1$ and $\alpha = 10$. To ensure a fair comparison, we conduct additional simulations under the adapted experimental setup as the real-robot demonstrations for the second experiment.
\subsection{Performance metrics} \label{performance}

We show the performance of our DAO-governed robot swarm using the following metrics: 
(i)~The \emph{average absolute error} defined as the absolute difference between the swarm's estimate and the true fraction of white tiles. This value is computed by taking, at each timestep, the swarm estimate for each run (the average of individual robot estimates as recorded in the smart contract of the robot with longest blockchain), computing its absolute error from the actual fraction of tiles (25\%), and then averaging these absolute errors across all runs to quantify the swarm’s collective sensing performance for an experimental set. (ii)~The \emph{true positives} which represent identification accuracy, i.e., the proportion of Byzantine robots that were removed by the end of a run, computed as the number of removed Byzantine robots divided by the total number of Byzantine robots. (iii)~The \emph{false positives} which represent identification error, i.e., the proportion of honest robots that were removed by the end of a run, computed as the number of removed non-Byzantine robots divided by the total number of non-Byzantine robots. (iv)~The \emph{average number of robots} in the swarm over time across all the runs of an experiment. (v)~The \emph{average number of Byzantines} in the swarm over time across all the runs of an experiment. (vi)~The \emph{average number of robots removed} from the swarm across all the runs of an experiment.


\section{Data Availability} 
The code used to generate the simulated data and the robot controller is open source and publicly available in the following GitHub repository: \url{https://github.com/rainazakir/dao}

\bibliography{sample}


\section*{Funding Statement}

R.\,Zakir, V.\,Strobel and M.\,Dorigo acknowledge support from the Belgian F.R.S.-FNRS, of which they are a FRIA Doctoral student, Postdoc and a Research Director, respectively. A. Reina acknowledges support from DFG under Germany's Excellence Strategy -- EXC 2117 - 422037984.

\section*{Author contributions statement}

V.S. and A.R. conceived the original idea. V.S. directed the project. R.Z. designed and implemented the control code for the robot simulations and ran them. R.Z., V.S. and A.M.P. ran the robot demonstrations. R.Z. generated the figures. M.D. provided the computing and robotics infrastructure that supported the project. R.Z. wrote the first draft of the paper, and all authors reviewed  and revised the paper.

\section*{Competing interests statement}

The authors declare no competing interests.


\end{document}